\pdfoutput=1

\documentclass[11pt]{article}

\usepackage[final]{acl}

\usepackage{times}
\usepackage{latexsym}

\usepackage[T1]{fontenc}

\usepackage[utf8]{inputenc}

\usepackage{microtype}

\usepackage{inconsolata}

\usepackage{graphicx}

\usepackage{multirow}

\usepackage{latexsym}
\usepackage{amssymb}
\usepackage{amsmath}
\usepackage{amsthm}
\usepackage{booktabs}
\usepackage{enumitem}
\usepackage{graphicx}
\usepackage{color}

\usepackage{times}
\usepackage{soul}
\usepackage{url}
\usepackage{hyperref}
\usepackage[utf8]{inputenc}
\usepackage{caption}
\usepackage{algorithm}
\usepackage{algorithmic}
\usepackage[switch]{lineno}

\usepackage{bm}
\usepackage{bbm}
\usepackage{mathtools}
\usepackage{xcolor}
\usepackage{stmaryrd}
\usepackage{adjustbox}
\usepackage{subfig}
\usepackage{amsfonts}
\usepackage{xspace}

\usepackage[normalem]{ulem}

\newcommand{\newadd}[1]{#1}

\newcommand{\mnli}[0]{\textsc{Mnli}\xspace}

\newcommand{\boolq}[0]{\textsc{Boolq}\xspace}

\newcommand{\cnndm}[0]{\textsc{CNN-DM}\xspace}
\newcommand{\samsum}[0]{\textsc{SAMSum}\xspace}
\newcommand{\vax}[0]{\textsc{VAX-Stance}\xspace}
\newcommand{\caves}[0]{\textsc{Caves}\xspace}
\newcommand{\squad}[0]{\textsc{SQuaD-v2}\xspace}

\newcommand\blfootnote[1]{%
  \begingroup
  \renewcommand\thefootnote{}\footnote{#1}%
  \addtocounter{footnote}{-1}%
  \endgroup
}

\title{Towards Sustainable NLP: Insights from Benchmarking \\ Inference Energy in Large Language Models\textsuperscript{$\dagger$}}

\author{
 {Soham Poddar\textsuperscript{1}*},
 {Paramita Koley\textsuperscript{2}*},
 \textbf{Janardan Misra\textsuperscript{3}},
 \textbf{Sanjay Podder\textsuperscript{3}},\\ 
 \textbf{Niloy Ganguly\textsuperscript{1}}, 
 \textbf{Saptarshi Ghosh\textsuperscript{1}} \\
 \textsuperscript{1} Indian Institute of Technology, Kharagpur, India. \\
 \textsuperscript{2} Indian Statistical Institute, Kolkata, India. \\
 \textsuperscript{3} Accenture Labs, Bangalore, India. \\
}

\begin{document}
\maketitle
 \vspace*{-12mm}

\blfootnote{$\dagger$ This paper has been accepted at NAACL main 2025.}
\blfootnote{* Equal Contribution}

\begin{abstract}
Large language models (LLMs) are increasingly recognized for their exceptional generative capabilities and versatility across various tasks. However, the high inference costs associated with these models have not received adequate attention, particularly when compared to the focus on training costs in existing research. 
In response to this gap, our study conducts a comprehensive benchmarking of LLM inference energy across a wide range of NLP tasks, where we analyze the impact of different models, tasks, prompts, and system-related factors on inference energy.
Specifically, our experiments reveal several interesting insights, including strong correlation of inference energy with output token length and response time. Also, we find that quantization and optimal batch sizes, along with targeted prompt phrases, can significantly reduce energy usage.
This study is the first to thoroughly benchmark LLM inference across such a diverse range of aspects, providing insights and offering several recommendations for improving energy efficiency in model deployment.
\end{abstract}

\section{Introduction}

\begin{table*}[ht]
    \centering
    \footnotesize
    \begin{tabular}{p{26mm}|c|c|c|c|c|c|c|c}
    \toprule
    & Fine- & Top- & Task & Response  & Model size & Batch  & Quanti- & Targeted  \\
    & grained  I/O & level  I/O & Complexity & time & \& family  &  size & zation & phrase \\
    
    \midrule
 Our Work                     & Yes & Yes & Yes & Yes & Yes & Yes & Yes & Yes \\
\citet{luccioni2024power}      & No  & Yes & No & No  & Yes & No & No & No \\
\citet{li2024toward}           & No  & No & No & No  & No & No & No & Yes \\
\citet{everman2023evaluating}  & No  & No & No & No  & Yes & No & No & No \\
\citet{desislavov2021compute}  & No  & No & No & No  & Yes & No & No & No \\
\citet{samsi2023words}         & No  & No & No & No  & Yes & No & No & No \\

    \bottomrule
    \end{tabular}
    \caption{Comparing our approach with existing literature on energy for specific experimental settings}
    \label{tab:comparison}
\end{table*}

%



Recent discussions on the energy and carbon impact of machine learning (ML) algorithms have mainly concentrated on quantifying energy usage during the training phase of these models~\cite{dodge2022measuring, luccioni2023counting, patterson2021carbon, raffel2020exploring}. Studies on inference energy are much rarer because a single inference operation consumes considerably less energy and resources. However, under deployment, inference is performed many more times, making its energy impact significant and warranting separate investigation~\cite{wu2022sustainable,patterson2022carbon}. For example, 90\% of total cloud computing demand for AWS, the largest global cloud provider, were for model inference purpose~\cite{barr2019amazon}. 
%
Moreover, a key motivation for training large language models is that a single model can achieve state-of-the-art performance across diverse NLP tasks due to its impressive zero-shot and few-shot capabilities, making it energy-efficient from a training perspective. However, when we consider the total carbon footprint of the entire lifetime of the model, the energy requirement for model inference plays a significant role, considering the number of inferences that are carried out during the model's lifetime. 
Thus it is crucial to conduct a systematic study to quantify the energy requirements and carbon emissions for model inference across various models and tasks.

\subsection{Literature survey}  Existing works have attempted to study the inference energy of LLMs from various perspectives. \citet{everman2023evaluating} evaluated energy usage of GPT models on question-answering tasks, employing Software Carbon Intensity (SCI) released by green software. \citet{samsi2023words} provides a detailed account of energy usage during inference for question-answering tasks for Llama models on various GPU architectures. \citet{liu2022few} study the energy consumption trade-off for fine-tuning vs few-shot learning for both encoder-decoder and decoder-only models on SuperGLUE and RAFT tasks, measuring energy in FLOPS. These preliminary works mostly ignore the performance-energy tradeoff. 

In recent times, \citet{li2024toward} compare the effect of prompt directives on inference energy and performance for Llama-2 models for 3 question-answering tasks. \citet{luccioni2024power} offers a comparatively detailed account of inference energy usage while choosing LLMs and tasks from a diverse range, showing the dependency of energy with model size and architecture, task type, and output token length. While these works provide an initial account of inference energy usage of LLMs in different configurations, they are often limited by the number and diversity of the models and tasks (Check Appendix~\ref{sec:appen-literature} for details).

\subsection{Our contributions \& differences with prior works}
In this work, we present a comprehensive study of LLMs, running models from both encoder-decoder and decoder-only models on both discriminative and generative NLP tasks, while analyzing the impact of different models, tasks, prompts, and system-related factors on inference energy.
%
Specifically, 
(i)~We start with a detailed analysis of various model-related factors that affect the inference energy of LLMs, where we systematically study the correlation between inference energy and influencing factors like input and output token length, response time, model size and complexity. 
(ii)~We conduct various experiments to study the connection between inference energy and batch size, level of quantization, and prompt editing. 
(iii)~We then complement our analysis by introducing the Normalized Accuracy metric, providing an accuracy-energy usage tradeoff analysis across tasks and models. 
(iv)~Finally, we present a list of efficiency guidelines in Section~\ref{sec:conc}.
%



In particular, our contributions include an energy analysis from eight aspects, as given in Table~\ref{tab:comparison}. Out of these aspects, several were primarily unexplored in the literature, as per our knowledge - these are indicated in the table. 
We vary the input and output token length at a finer granular level while reporting the variation in energy to have a better understanding of the correlation between energy and these factors, revealing several valuable insights (Section~\ref{sec:input-output}). A more thorough analysis of energy with variation of input or output while keeping the other fixed provides a more accurate insight into their comparative influence in inference energy, implying immediate solutions for energy-efficient inference approaches. The correlation between energy and response time validates response time as a good proxy of inference energy in black-box models (Section~\ref{sec:responsetime}). Our work first explores task complexity's effect with inference energy (Section~\ref{sec:complexity}). We also provide the performance change for quantization (Section~\ref{sec:batch}) and use of targeted phrases (Section~\ref{sec:phrases}) along with improvement in energy. While the effects of quantization and increasing batch size have been studied individually, we show that using them together under fixed memory constraints can help speed up the computations. Finally, we explore each of the above aspects for a diverse range of LLMs from both architecture families along with a diverse range of tasks, which was mostly lacking in previous attempts.

\section{Experimental setup}

In this section, we describe the models, datasets and various settings we use for our experiments.


\subsection{Models}
\label{sub:models}

We select $6$ popular and recent GPT-style models from the decoder-only family and $4$ Flan-T5 models from the encoder-decoder family, adding to $10$ models in total 
(details in Appendix~\ref{app:models}).

\noindent \textbf{Decoder-only Models} generate output in an autoregressive manner by predicting the next token in the sequence based on the context (key-value-query) vectors corresponding to the input and previously generated tokens. 
We consider the following models from decoder family in our study. 
(D1)~\textbf{Tiny-LLama}~(1.1B params);
(D2)~\textbf{Phi-3-mini}~(3.8B params);
(D3)~\textbf{Mistral-7B}~(7.2B params);
(D4)~\textbf{Llama-2-7B}~(6.7B params);
(D5)~\textbf{Llama-3-8B}~(8.0B params);
(D6)~\textbf{Llama-2-13B}~(13B params);

\noindent \textbf{Encoder-Decoder models} process the input data and convert it into context (key-value) vectors. Then the decoder takes these vectors and generates output autoregressively. Models from this family, considered in our study, include: 
(ED1)~\textbf{Flan-T5-base}~(248M params); 
(ED2)~\textbf{Flan-T5-large}~(783M params);
(ED3)~\textbf{Flan-T5-xl}~(2.8B params);
(ED4)~\textbf{Flan-T5-xxl}~(11B params);


\subsection{Tasks and Datasets} 
In this work, we select a diverse range of NLP tasks, from generative to question-answering, classification, and single-sentence tasks. This includes both general GLUE / SuperGLUE benchmarks, as well as domain specific \vax and \caves (for studying effect of task complexity).
We describe the tasks and their corresponding datasets in Table~\ref{tab:tasks}. 
%
%
%
%
%
%
For each dataset, we selected $1024$ data samples randomly and performed all experiments on the same set for comparable results.
Performance metrics are chosen depending on the tasks. For summarization tasks, average of ROUGE1, ROUGE2, and ROUGE-L are reported, whereas some form of F1 score are reported for the other tasks. Description/prompts of the datasets and the individual metrics have been given in Appendix~\ref{app:tasks}.

\vspace{1mm}
\noindent \textbf{Normalized Accuracy (NA) Metric:}
Since different tasks use different metrics on different scales, it is difficult to compare the accuracy performance of models across the tasks. To gauge the overall performance of the models across multiple tasks, we introduce the \textit{NA} metric that is obtained as follows. For each dataset, we first perform Z-score normalization across all the models, followed by a sigmoid operation to scale models between $0$ and $1$. We then average the scores for each model across all datasets and multiply by $100$.
Note that this metric depends on the set of models used and will vary if models are added/removed. However, it allows us to quantify how well a model performs compared to others in the set.

\begin{table}[!t]
    \centering
    \footnotesize
    \begin{tabular}{p{40mm}|p{28.5mm}}
    \toprule
    \textbf{Task} &
    \textbf{Dataset} \\
    \midrule
    Linguistic acceptability check & COLA (GLUE) \\
    Logical entailment & Mnli (GLUE)  \\
    Sentiment classification & SST2 (GLUE)  \\
    \hline
    Contextual question answering & Boolq (SuperGLUE)  \\
    Causal reasoning & COPA (SuperGLUE)  \\
    Entity Question answering & ReCoRD (SuperGLUE)  \\
    Extractive question answering & SQuAD v2 \cite{rajpurkar-etal-2016-squad} \\
    \hline
    Document summary generation & CNN-DM \cite{nallapati2016abstractive} \\
    Dialogue summary generation & SAMSum \cite{gliwa2019samsum} \\
    \hline
    3 class vaccine-stance classification & VAX-Stance \cite{poddar2022winds} \\
    12 class multi-label anti-vaccine concerns classification & CAVES \cite{poddar2022caves} \\
    \bottomrule
    \end{tabular}
    \caption{List of tasks/datasets we experimented on. Most tasks are taken from the GLUE~\cite{wangglue} and SuperGLUE\cite{wang2019superglue} benchmarks. Description/prompts are been given in Appendix~\ref{app:tasks}.}
    \label{tab:tasks}
\end{table}

\subsection{Hardware and Energy metrics}
We perform our experiments on a single NVIDIA A6000 GPU with 48GB VRAM hosted in a local server with Intel Xeon Silver 4210R processor and 128GB RAM, running Ubuntu 20.04-LTS. 
The server also hosted an NVIDIA A5000 GPU (24 GB), which was used for only one experiment, but otherwise was unused.
We also performed some experiments on two other systems to verify the generalizability of our findings, as detailed in Appendix~\ref{app:mgpu}.
We use Pytorch version 2.3 (with CUDA 12.1) and huggingface transformers version 4.41.

We use the popular Code Carbon~\cite{schmidt2021codecarbon} and Carbon Tracker~\cite{anthony2020carbontracker} packages to measure the energy consumed in different experiments. 
\citet{jay2023experimental} demonstrated the suitability and accuracy of CarbonTracker, CodeCarbon, Energy Scope, and Experiment Impact Tracker across various software-based power meter setups, while \citet{bouza2023estimate} further established the superiority of CodeCarbon and CarbonTracker among these tools. CodeCarbon is especially the most user-friendly and works out of the box, provided appropriate NVIDIA libraries and permissions to Intel RAPL files.

%
These two packages measure the GPU-power usage using pynvml and CPU-power using Intel RAPL files every $\mathcal{X}$ seconds, and integrates it over time. Carbon-tracker reports sum of these as the total energy.
Code-carbon also adds an estimate of the RAM-power being used depending on the RAM size. 
We use $\mathcal{X} = 10secs$ for a balance between overhead costs and tracking accuracy.
We keep the Power Usage Effectiveness~(PUE) to the default $1.0$ since we run all experiments on the same server, but this implies that the actual energy usage is higher than reported.

During inference, we provide test samples in batches to the LLM, 
and measure the total energy required for 1024 samples per dataset using these tools. This includes both the input tokenization process by each model's tokenizer and the output generation from the model.
We keep the batch size to $8$ for most experiments, except on the \cnndm and \samsum dataset for which we use a batch size of $4$. 
While reporting results, we average the energy usage and report the \textbf{energy per sample} in mWh (milli-Watt-hour). Unless otherwise stated, these are the default settings used for experiments.


\section{Factors Affecting Energy / Accuracy}
In this section, we discuss how different task, model, and setup-related factors contribute to the inference energy and accuracy metrics.

\begin{figure*}[!t]
\centering

\includegraphics[height=4.25mm,trim={85mm 141mm 70mm 2mm},clip]{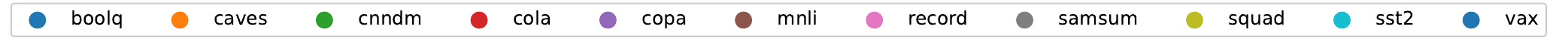}
\vspace*{-6mm}

\subfloat{\includegraphics[height=4cm,trim={3.5mm 4mm 3.5mm 3.5mm},clip]{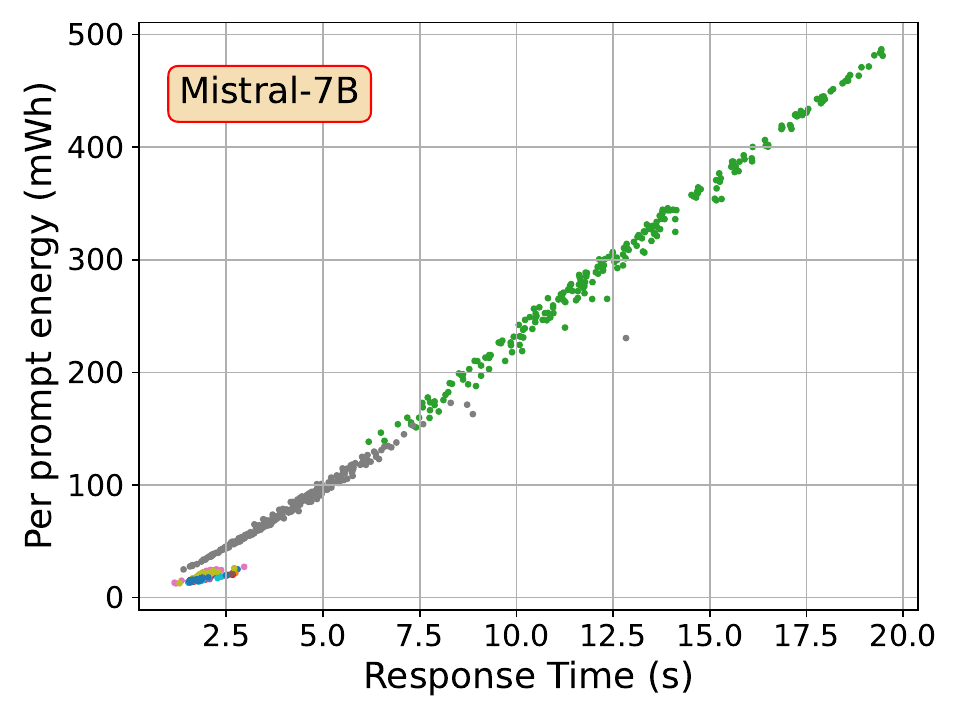}}
\hfill
\subfloat{\includegraphics[height=4cm,trim={10.5mm 4mm 3.5mm 3.5mm},clip]{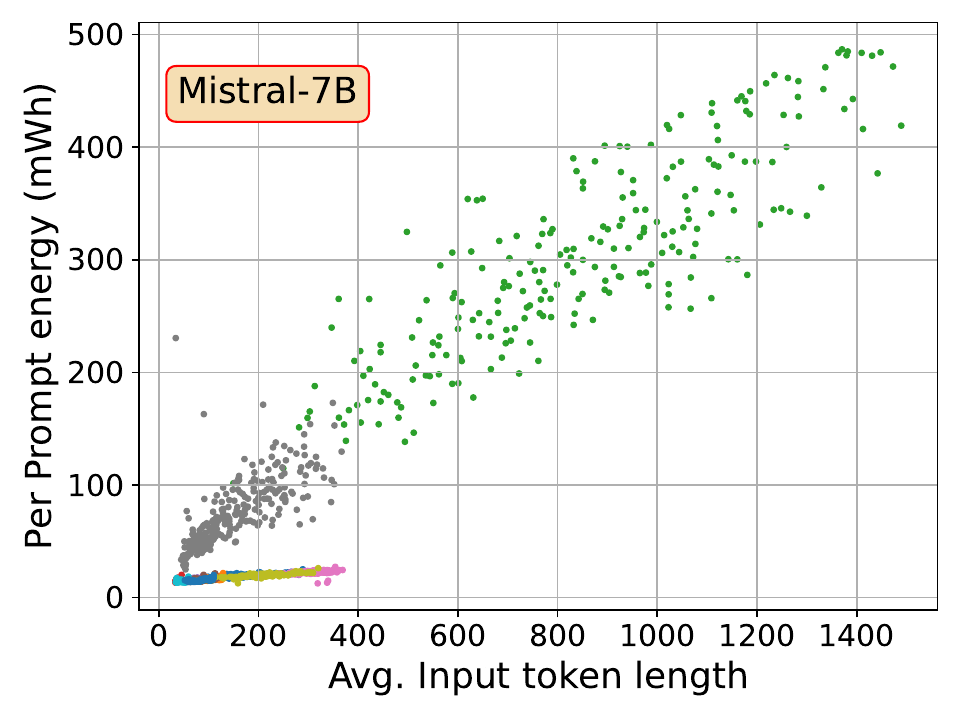}}
\hfill
\subfloat{\includegraphics[height=4cm,trim={10.5mm 4mm 3.5mm 3.5mm},clip]{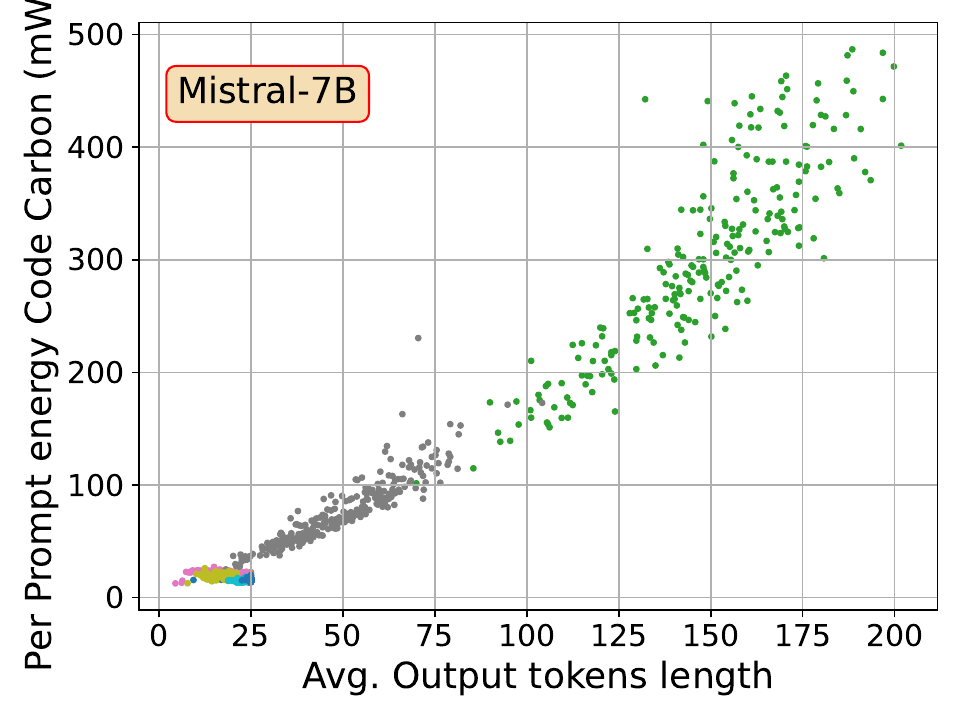}}

\caption{Inference energy vs response time, input and output-token length averaged across samples in a batch plotted across all datasets for  \textbf{Mistral-7B}. Dots correspond to distinct batches of different datasets.}
\label{fig:energy-scatter}
\end{figure*}

\subsection{Response time}
\label{sec:responsetime}
Response time is an important indicator for actual inference energy, as given a (somewhat) fixed amount of power draw, the energy consumed is proportional to inference response time. 
To this end, we track the energy for each batch where the tracking interval is set to $1 sec$ for the energy-measuring libraries. The batches were formed after sorting the inputs (prompt + query) by length (so that similar-length queries end up together, allowing optimal padding and energy usage). 

Figure~\ref{fig:energy-scatter} (left column)
compares per-sample average response time and inference energy.
They report the comparison for Mistral (rest of the models in Appendix~\ref{app:scatter}). 
Points in the plot correspond to the average scores per query for the individual batches, with distinct color for each dataset. 

We find a strong correlation between response time and the inference energy (pearson $r = 0.996$, spearman $rs = 0.968$), indicating a strong possibility of using the response time as a reliable proxy for the energy consumed if demographic factors like location, energy grid, model, etc, are fixed.
However, for different datasets, the slope of the dependency is different, which may be because of slightly different power draws due to datasets having different-sized inputs.
We also compare the energy measures returned by CarbonTracker and CodeCarbon package and find a good correlation (pearson $r = 0.610$, spearman $rs = 0.912$), indicating reliable tracking.

\subsection{Input and Output token length}
\label{sec:input-output}

The complexity of each attention block in a transformer decoder model is given by the following equation ~\cite{vaswani2017attention}, where $n$ is the \#input~tokens, $d$ is the hidden dimension, and $t$ is the \#output tokens.
%
\begin{equation}
\label{eq:llm-complexity}
O(n,d,t) = (n.d^2+n^2.d).t
\end{equation}

This equation suggests input and output length play a major role in deciding the computational complexity of large language models, which consist of several consecutive layers of multiple such attention blocks, and thereby, the required inference energy. 
In this section, we attempt to explore the influence of the aforementioned factors in a more systematic manner. 
Toward that, 
we first explore a similar setup explained in Section~\ref{sec:responsetime} to plot the batch-wise energy. Sorting the inputs by their length before batching is especially important because the batches with random input lengths can all get averaged out to have similar values otherwise, making it difficult to visualize the effect of energy with input/output sizes.
%


\vspace{2mm}
\noindent \textbf{Input Length:}
Figure~\ref{fig:energy-scatter} (middle column)
compares per-sample average inference energy with per-sample average input token length.  
The bigger, spread out clusters primarily belong to the generative tasks, namely, \cnndm{} and \samsum{} datasets, due to their larger outputs than the other discriminative tasks, which lay in the bottom clusters. 
Even though the input size appears as a quadratic term in Eq.~\ref{eq:llm-complexity}, we see a linear variation of energy usage with input size.
\newadd{This discrepancy can be attributed to various factors -- hardware-related optimizations such as parallel processing, caching mechanisms, and memory hierarchies – that significantly influence the relationship between computational complexity and energy usage.
For instance, Transformer models (including LLMs) leverage GPU parallelism to process the input, which effectively mitigates the quadratic scaling predicted by theoretical calculations~\cite{vaswani2017attention}. Additionally, mechanisms like key-value (KV) caching ensure that input tokens are processed only once, further reducing energy demands compared to the theoretical worst-case scenario~\cite{pope2022efficiently}. These optimizations are likely the primary contributors to the observed linear correlation between input token length and energy consumption. }
%

\vspace{2mm}
\noindent \textbf{Output Length:}
Figure~\ref{fig:energy-scatter} (right column)
compares per-sample average output token length with per-sample average inference energy for individual batches. Here we observe a similar trend indicating linear increment of energy with output size, in accordance with Eq.~\ref{eq:llm-complexity}. 
Here most tasks except \cnndm{} and \samsum{} cluster around the bottom left because of their short outputs, whereas the widespread clusters of \cnndm{} and \samsum{} towards the top provide a better visualization of the linear dependency.
\newadd{Note that, the slope of variation of energy consumption is steeper for the output length, in comparison to the slope for the input length despite the input being larger than the output (input Pearson $r = 0.697$, output Pearson $r = 0.952$). These observations indicate \textit{a stronger role played by output length (than the input length) in deciding the inference energy}, which can be explained as follows. 
During inference, LLMs leverage key-value (KV) caching, which allows tokens to be processed only once,
avoiding repeated prior token processing. Furthermore, input processing can be parallelized on GPUs, leading to significant speedups and lower energy consumption~\cite{vaswani2017attention}. 
In contrast, generating output tokens is inherently a sequential process, as each token depends on the previous one. This lack of parallelism for output token generation leads to the steeper slope of output token length. 
}

\begin{figure}[!t]
\centering
{\includegraphics[height=50mm,trim={4mm 4mm 3mm 4mm},clip]{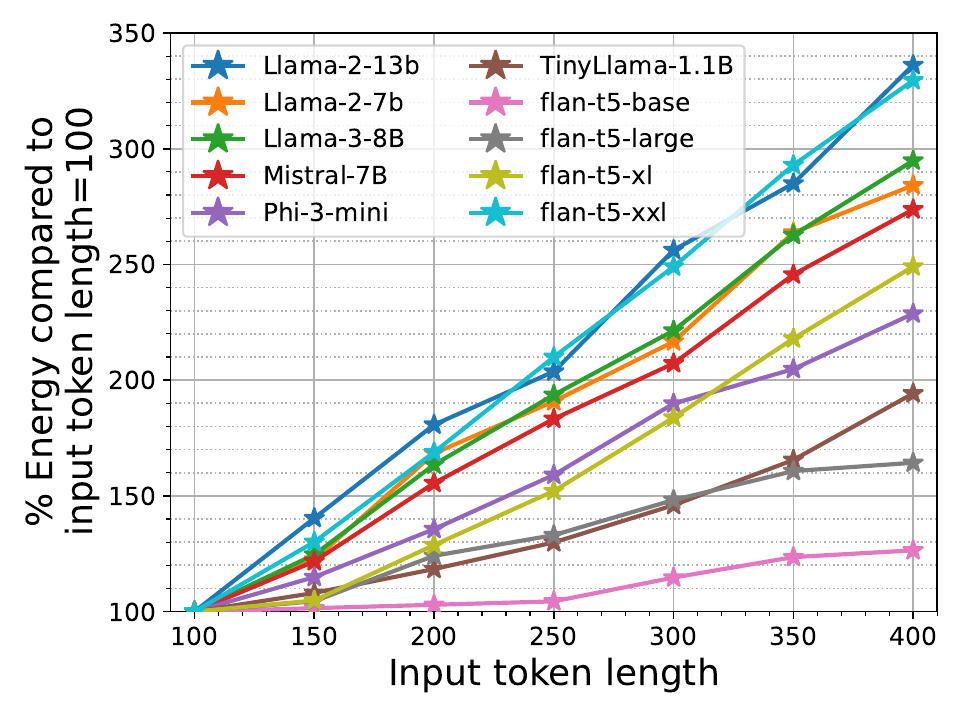}}
\caption{Inference energy on \cnndm where we vary input token lengths fixing \#output tokens to 1.}
\label{fig:vary-input}
\end{figure}

\vspace{2mm}
\noindent \textbf{Controlled setup:}
For better and more exclusive insights into the relation between inference time and input and output length, we perform the following controlled experiment where we fix either input or output length and vary the other.

\noindent \textbf{\textit{Effect of varying input length}:}
For \cnndm dataset, we truncated each input text into $N$ tokens and ask the model to summarize the input, where we vary $N$ from $100$ to $400$ at fixed intervals of $50$ tokens, by means of truncation/padding. If input has more than N token,  
the query text is truncated, and if the input contains less than N tokens, then padding tokens are added to the left of the input.
To eliminate the influence of generated output on inference energy, we stop generation after the first token, which allows us to monitor the influence of input length on model inference energy in an exclusive manner.

Figure~\ref{fig:vary-input} plots the inference energy (in terms of \%) relative to the energy required for the input with $100$ tokens. The results indicate a linear increase in energy with longer input lengths, with a steeper slope observed for decoder-only models. This is likely due to the longer decoder modules present in these models.

\begin{figure}[!t]
\centering
\includegraphics[height=50mm,trim={4mm 4mm 3mm 4mm},clip]{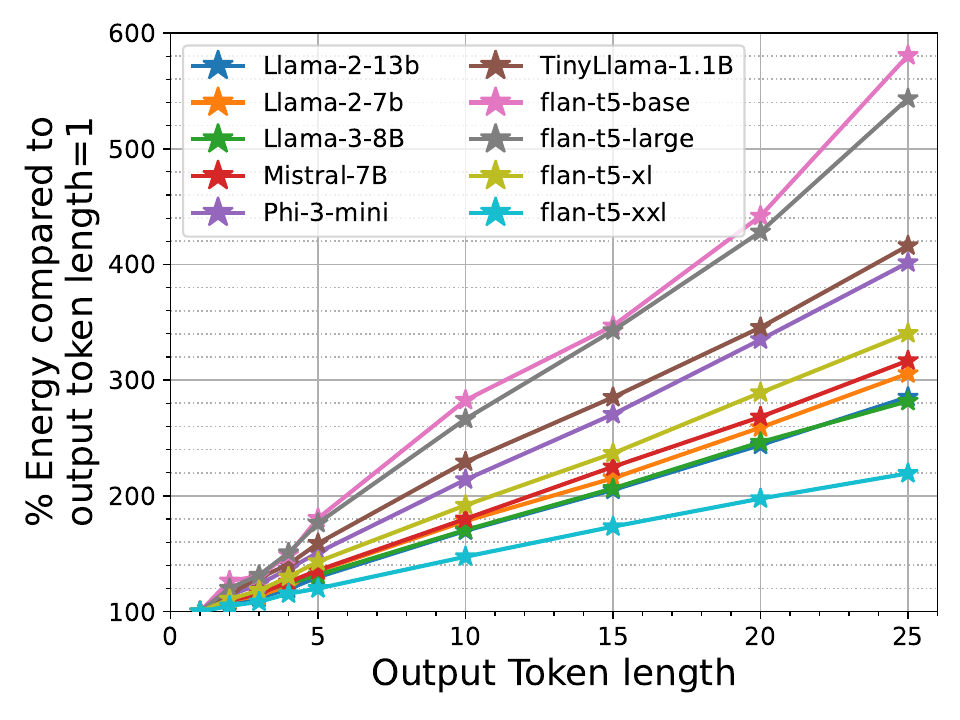}
\caption{Inference energy on \cnndm{} dataset when the output length is varied, keeping input length fixed. }
\label{fig:vary-output}
\end{figure}

\noindent \textbf{\textit{Effect of varying output length}:}
Similarly, to study the effect of output length on inference energy exclusively, we take the \cnndm dataset, fix the input text length and allow the generation length to vary from $1$ to $25$ tokens. Specifically, we instruct the model to summarize the input text 
but force the model to stop after it generates the required number of tokens. 

Figure~\ref{fig:vary-output} plots the energy (in \%) relative to the energy required for generating a single token, confirming linear energy increment with generation length.
However, the energy of generating the $1^{st}$ token is much more than additional tokens, e.g. generating $2$ tokens takes only about 12\% more energy than generating $1$ token. This is because the model processes the entire input in the first time step, but only 1 token for subsequent steps (by means of caching the K-V computations for prior tokens).
Also, note that, the increment of energy is larger with increasing output length, compared to input length.  
Contrary to Figure~\ref{fig:vary-input}, the relative increase is higher for the encoder-decoder family here, attributed to the fact that initial energy requirement is smaller for these families, along with higher jumps in energy with output length. 
%

\subsection{Task complexity}
\label{sec:complexity}

Next, we explore whether task `complexity' has a significant impact on inference energy. Toward that, we conduct a series of controlled experiments 
where inference energy of two tasks with identical input and output length and distinctly different levels of complexity are compared. 
Here, we interpret "complexity" based on human cognition.

We compare the inference energy between the \vax and \caves datasets, where the input texts are similar—tweets related to vaccines—but the tasks differ: a 3-class single-label classification for \vax{} versus a 12-class multi-label classification for \caves{}.
\newadd{
We ensure consistent input length via padding and fix the output to a single token.
We find 
the difference in energy consumption between the two tasks to be very small ($<1\%$), as shown in Table~\ref{tab:task_complexity}.
}

Similarly, we compare the average inference energy for the summarization (hard) vs returning the first three sentences of the input (trivial) over the \cnndm dataset. 
We find the energy difference between the two tasks as less than 1.3\% (Table~\ref{tab:task_complexity}), again indicating that task complexity has hardly any impact on inference energy if input and output lengths are kept fixed.  
This observation follows from the fact that the computational steps per token are fixed by the model's architecture, with  
LLMs processing the inputs uniformly, without additional branches or conditional logic that would increase the load for more complex tasks.

\newadd{
Although we explore task complexity as perceived by humans, results could be different in more complex Chain of Thought~\cite{wei2022chain,wang2022towards}, Tree of Thought~\cite{yao2023tree} and other reasoning frameworks. 
These frameworks have different ways of processing a task, which could impact the energy consumption based on the settings used. 
Comparing energy consumption for these different settings would need a study of its own, and is left as future work.
}

\begin{table}[t]
    \centering
    \footnotesize
    \begin{tabular}{|l|c|c|}
    \hline
               & \textbf{CAVES vs} & \textbf{Summarization vs} \\
               &              \textbf{VAX-Stance} & \textbf{first 3 sent extraction} \\
    \hline
    fT5-large  & $-0.2$\% &  1.3\% \\
    fT5-xl     &  0.5\% & $-0.6$\% \\
    Phi3-mini  &  0.4\% & $-0.1$\% \\
    Mistral-7B &  0.4\% & $-0.3$\% \\
    Llama3-8B  &  0.5\% &  0.8\% \\
    \hline
    \end{tabular}
    \caption{Percentage difference in energy consumption for two tasks with very different complexities, keeping input and output lengths same. The differences are very small.}
    \label{tab:task_complexity}
\end{table}

\subsection{Model family and size}
\label{sec:model-family}

\begin{figure}[!t]
    \centering
    \includegraphics[width=\linewidth,trim={4mm 9mm 3mm 3mm},clip]{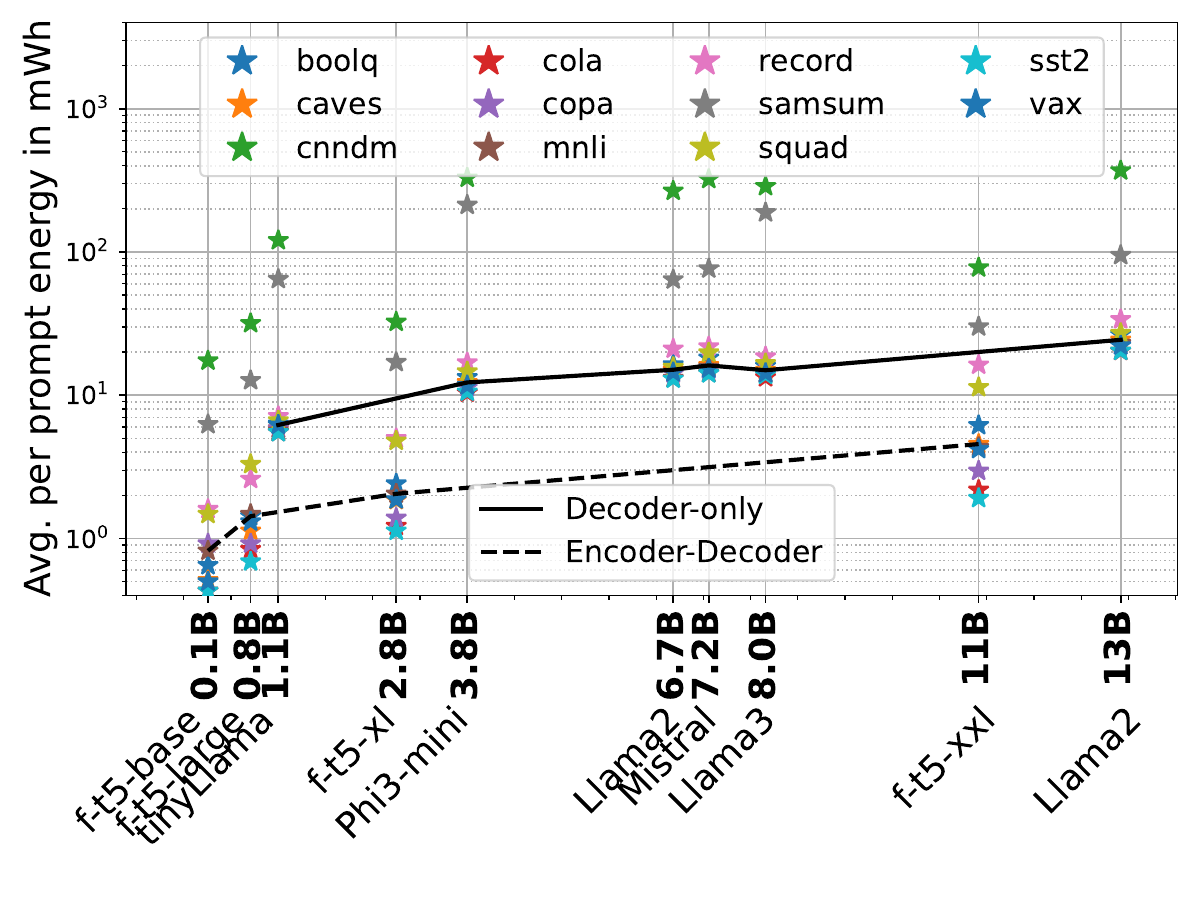}
    \caption{Average per-prompt inference energy vs model size for all models and datasets. The black lines join the median energy for each model family.}
    \label{fig:model_size}
\end{figure}

We now compare the energy usage and normalized accuracy of different models with respect to their size (number of parameters). The model sizes and family have been listed in Section~\ref{sub:models}. 
Figure~\ref{fig:model_size} 
compares the size of models with per-sample inference energy, averaged across all samples, showing 
a linear increase with the size of the model 
(note that only Y-axis is in log scale in Fig~\ref{fig:model_size}), that are
individually visible for both the encoder-decoder and the decoder-only models.



Encoder-decoder models typically consume less energy than decoder-only models with a comparable number of parameters. 
For instance, Flan-T5-xl and Phi-3-mini have a similar parameter count but use significantly less energy. 
The same pattern holds true for Flan-T5-large versus tinyLlama, and Flan-T5-xxl versus Llama-2-13B. 
This is because the decoder part in the encoder-decoder models is half the size, which reduces computational demands during the autoregressive decoding phase.

%
%
%

\begin{table*}[!t]
    \begin{center}
    \footnotesize
    
        \begin{tabular}{|c||c|c|c|c||c|c|c|c|c|c|}
        \hline
        \ & flan-t5 & flan-t5 & flan-t5 & flan-t5 & TinyLlama & Phi-3 & Mistral & Llama-2 &  Llama-3 & Llama-2 \\
                                & base    & large    & xl      & xxl    & 1.1B      & mini  & 7B      & 7B      & 8B & 13B\\
        \hline \hline
        \multicolumn{11}{|c|}{\it \textbf{I.} Average \textbf{Normalized Accuracy} across all datasets with original settings} \\
        \hline
        default & 42.85 & 69.77 & 55.45 & 58.12 & 23.5 & 41.82 & 59.91 & 42.83 & 55.09 & 52.31\\ 
        \hline \hline

        \multicolumn{11}{|c|}{\it \textbf{II.} Average \textbf{change in performance (\%)} on Quantization} \\
        \hline
        8-bit  & 0.47 & 0.03 & -1.16 & 1.43 & 0.9 & -2.92 & -0.55 & -0.46 & -2.43 & -4.66 \\  
        4-bit  & 1.78 & -1.83 & -0.63 & -0.61 & 9.66 & -4.19 & 4.27 & -1.57 & -1.95 & -2.76\\ 
        \hline \hline
        
        \multicolumn{11}{|c|}{\it \textbf{III.} Average \textbf{change in performance (\%)} on introducing targeted phrases in prompts} \\
        \hline
        fix-output & -1.7 & -2.42 & -1.22 & 0.39 & 4.71 & -8.21 & 11.64 & -12.54 & -8.53 & 3.24\\
        \hline
        
        energy-eff & -1.28 & -4.2 & -0.89 & 0.84 & 1.44 & 9.55 & 2.52 & 0.33 & -5.19 & 5.2\\
        
         + fix-output & -1.33 & -2.68 & -1.78 & 0.74 & 4.07 & 4.11 & 12.4 & -15.73 & -10.07 & 5.38\\
        \hline
        quick & 1.29 & -3.88 & -0.41 & -1.71 & 2.63 & 8.14 & 5.82 & -4.24 & -14.88 & 3.69\\
        
        + fix-output & -0.72 & -5.48 & -0.27 & 0.2 & 4.64 & -2.55 & 12.45 & -13.23 & -18.88 & 2.64\\
        \hline
        \end{tabular}
        \caption{Accuracy metrics for LLM inferences averaged across all datasets. 
        \textbf{I.}~Encoder-Decoder models perform better or close to Decoder only models.
        \textbf{II.}~Quantization does not decrease performance by much ($<5\%$)
        \textbf{III.}~Performance degrades with most phrases, more so where energy / output token length had also reduced. 
        }
        \label{tab:metrics}
    \end{center}
\end{table*}

\vspace{1mm}
\noindent \textbf{Accuracy metrics:} The top row of Table~\ref{tab:metrics} lists the Normalized accuracy (NA) scores for each model across all datasets (performance on each dataset given in Appendix~\ref{app:scores}). 
Here, we observe that performance depends on both model size and family. Smaller models perform poorly, with TinyLlama giving the worst performance, followed by Phi-3-mini and flan-t5-base. Llama models tend to perform inferior to flan-t5 models of similar size (flan-t5-large, -xl, and -XXL). Mistral-7B is the only exception among the decoder-only models that performs comparably with the flan-t5 family. 

As a general statement, it can be commented that selecting models from the encoder-decoder family for NLP tasks is recommended from an energy-efficiency perspective, as well as their performance which is improved by sequence to sequence instruction tuning. 
In contrast, decoder-only models trained on vast amounts of general data is more suited as an informational chatbot (though instruction tuned versions of Llama3, Mistral and Phi-3 try to bridge the gap).


\subsection{Batch size and Quantization}
\label{sec:batch}
We try to understand the effect of batch size on the energy usage during inference. 
Intuitively, increasing the batch size should require more energy per batch but we show that it
requires less energy per individual sample.

\begin{figure}[!t]
\centering
{\includegraphics[height=50mm,trim={4mm 5mm 3mm 3mm},clip]{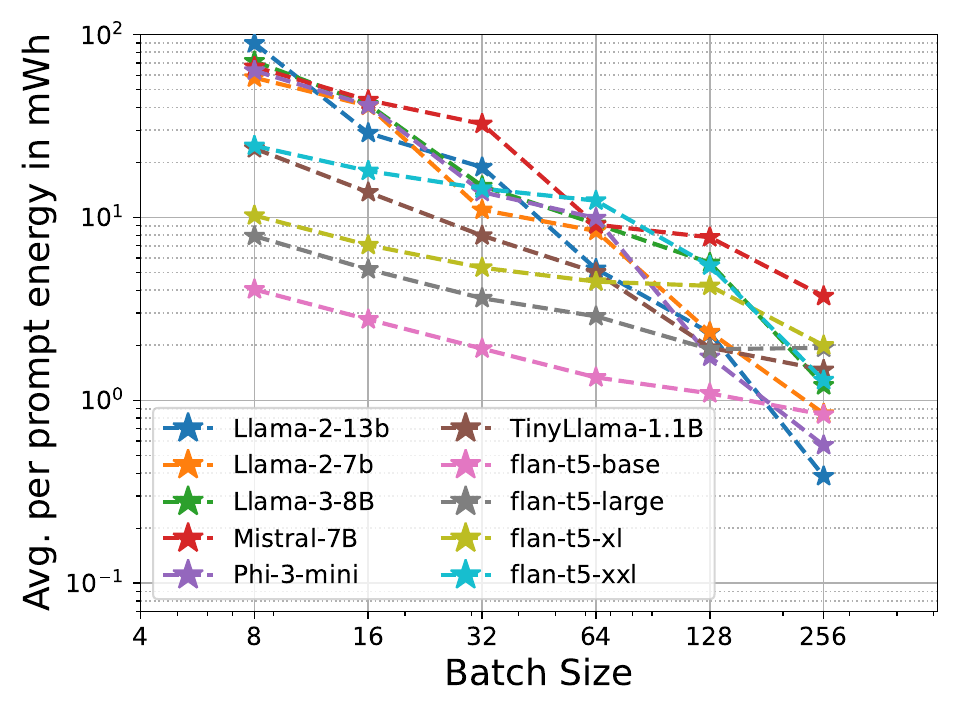}\label{fig:vary-batch-quantized}}
\caption{Per-sample inference energy with 4-bit quantized models when the batch size is varied, averaged across all datasets. 
}
\label{fig:vary-batch-quant4}
\end{figure}

We have used the \textit{bitsandbytes} package to load the transformers model weights in 8-bit and 4-bit quantized format. These quantized versions take up much less GPU memory to load (and thus can be run with larger batch sizes), though the computations still get executed in 16-bit single precision format. 
We run 4-bit quantized models on all the tasks under different batch sizes and plotted the average energy consumption per sample across all tasks in Figure~\ref{fig:vary-batch-quant4}. 
%
We observe that increasing the batch size leads to a decrease in per-sample inference energy.
%
%

However, the maximum batch size possible is limited by size of the GPU VRAM (48GB for A6000).
For certain datasets and larger model combinations, higher batch sizes can result in out-of-memory errors, suggesting that there is an optimal batch size for each dataset and model size combination. To achieve energy-efficient inferences, it is advisable to perform inferences close to this optimal batch size.%

%

\vspace{2mm}
\noindent \textbf{A5000 GPU:}
We repeated this experiment on an NVIDIA A5000 GPU instead of the A6000, reported in Appendix~\ref{app:bs_quant}; however, we did not find a significant difference in the inference time. 
This also signifies that GPUs with similar usable power require similar energy. Instead, the GPU VRAM size plays a more important role, allowing larger batches.


\vspace{2mm}
\noindent \textbf{Quantization Energy:}
%
%
%
Figure~\ref{fig:vary-batch-quant4} shows the average change in energy for the 4-bit quantized model, while the energy required by original model is given in the Appendix~\ref{app:bs_quant}. Interestingly, keeping all factors same, quantization increases the energy used to almost $2\times$, because of the overhead of additional data format conversions to 16-bit. However, using the 4-bit quantized model with larger batch size of 256 reduces the energy to about $0.33\times$ of the original 16-bit model with batch size of 64.
We noticed very similar results with 8-bit quantization and thus, its energy plot is given in Appendix~\ref{app:bs_quant}.

\vspace{2mm}
\noindent \textbf{Quantization Accuracy metrics:} the change in performance of the quantized models compared to the original is given in the middle set of rows of Table~\ref{tab:metrics}. Quantization seems to reduce the performance by less than 5\% for some models (mostly decoder-only models in 8-bit quantization and most of the models for 4-bit quantization) and even increases performance slightly for some smaller models, which may have been overfitting earlier.
Thus, quantization does not seem to degrade performance too much and should be used to speed up inference time by increasing batch size.

\subsection{Effect of targeted phrases in prompts on inference energy}
\label{sec:phrases}

Finally, we attempt to find whether addition of phrases targeted towards energy optimization can affect the inference energy. Specifically, we wanted to see if adding a few more input tokens can lead to a larger decrease in energy by reducing the output token length.
In this experiment, for each dataset, we append certain targeted phrases after the default prompt, as shown in Table~\ref{tab:prompts}.
%
%

\begin{figure}[!t]
\centering
{\includegraphics[width=\linewidth,trim={4mm 5mm 3mm 3mm},clip]{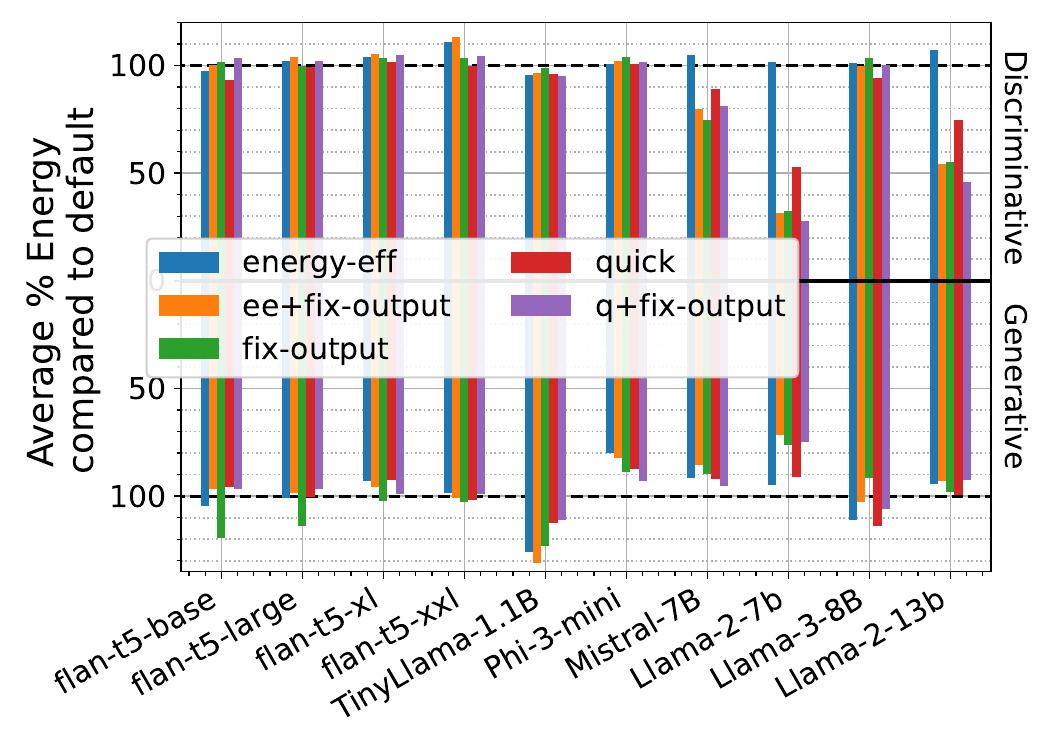}\label{fig:vary-prompt-energy}}
\caption{Effect of inserting targeted phrases in prompt on inference energy, as a percentage of default prompt. 'ee': energy-efficient, 'q': quick (see Table~\ref{tab:prompts}). }
\label{fig:vary-prompt-averages}
\end{figure}

Figure~\ref{fig:vary-prompt-averages} reports the \% inference energy usage using modified prompts compared to default prompts, averaged across the two generative and rest discriminative tasks. 
Significant energy reduction is observed for Mistral and Llama-2 models. 
The reduction is less pronounced in generative tasks, where it mainly results from slightly shorter outputs. However, for discriminative tasks, the reductions are much more significant. This difference arises because these models typically include explanations with their answers, leading to longer outputs by default. By instructing the model to be concise 
we can limit the output length and, thus, reduce inference energy.
However, the change in energy is negligible for the encoder-decoder models, Llama-3 and Phi-3-mini, as they typically generate short, brief answers, leaving little scope for reducing the output. Thereby, additional phrases in the prompt increase the input without reducing the generation, resulting in higher inference energy. 
TinyLlama always generates long outputs, often stopping only at the generation limit, rendering the targeted phrases useless.

%

\begin{table}[!t]
    \centering
    \footnotesize
    \begin{tabular}{p{14mm}|p{54mm}}
    \toprule
    \textbf{Directive} &
    \textbf{Targeted Phrases} \\
    \midrule
    default & Read the passage and answer the question with True or False. \\
    quick & \textit{<default>} Answer as quickly as possible. \\
    fix-output & \textit{<default>} Do not output anything else. \\
    energy-eff & \textit{<default>} Answer in energy-efficient way. \\
    \bottomrule
    \end{tabular}
    \caption{List of targeted phrases that are used to instruct the LLM for energy-efficient inference.}
    \label{tab:prompts}
\end{table}

\vspace{1mm}
\noindent \textbf{Accuracy metrics:} 
The performance metrics for modified prompts are given in bottom rows of Table~\ref{tab:metrics}, we observe that the introduction of such phrases in prompts results in diverse behavior depending on model size and architecture. 
Performance degrades in most of the cases, especially where output token length had also reduced, taking lesser energy.
In summary, we can see that the introduction of such phrases turns out to be useful only for Mistral-7B and Llama-2-13B, considering energy efficiency without affecting performance.







\section{Concluding discussions}
\label{sec:conc}

In this study, we benchmarked the power consumption of various large language models (LLMs) during inference across diverse NLP tasks.

\vspace{2mm}
\noindent \textbf{General implications:} 
Our primary high-level takeaways can be summarized as follows: 
\textbf{(1)}~%
\newadd{
Inference energy displays a strong correlation with response time for locally run open-source LLMs, making it a reliable proxy for estimating energy consumption, with significantly less overhead compared to using specialized energy measurement tools.
In the case of models accessed through online APIs (such as closed-source models), 
in the absence of energy-related metrics from the API providers, response time can play as a proxy for energy estimation (see Appendix~\ref{app:blackbox}). 
}
\textbf{(2)}~While input size shows a linear relationship with energy use, output length has a stronger influence on inference energy.
\textbf{(3)}~Task complexity has little impact on inference time independent of input and output lengths.
\textbf{(4)}~ 
Selecting models from the encoder-decoder family for NLP tasks is recommended from an energy-efficiency perspective, as well as their performance.
\textbf{(5)}~Increasing batch size reduces inference energy. However, it is constrained by the GPU memory availability, recommending an optimal batch size for a particular model, task pair. 
\textbf{(6)}~Quantization allows larger batches, resulting in lower energy use without degrading the inference accuracy much.
%
\textbf{(7)}~Introducing targeted phrases achieves energy reduction for older decoder-only models by restricting their output for discriminative tasks.

\vspace{2mm}
\noindent \textbf{Implications in energy-constrained environments:} In situations where LLMs are to be deployed in resource/energy constrained settings, our experiments lead to the following insights: 
\textbf{(i)}~The response time can be used as a proxy for energy consumed. This is useful not only since measuring power/energy may be difficult on most hardware but also because measuring response time requires much less overhead (in terms of energy/latency) than actually measuring the energy consumption. In low resource settings, this overhead can be large in terms of percentage. For example, we found that removing the energy-tracking libraries and just recording the system time can reduce inference time between 15\%--50\% under different scenarios.
\textbf{(ii)}~Where possible, input compression and output optimization should be employed. At least, some targeted phrases should be incorporated so that the model generates only what is needed.
\textbf{(iii)}~Fine-tuned encoder-decoder models (especially Flan-T5 models), are better suited for low-resource settings, particularly for NLP tasks such as classification, summarization, etc. 
\textbf{(iv)}~Quantized models should be employed, allowing larger batches and larger models on such settings, from the point of memory constraint.  

\vspace{2mm}



\section*{Limitations}
Despite the comprehensive analysis and valuable insights provided by this study, the following limitations should be considered. First, the benchmarking experiments were conducted under controlled conditions, which may not fully capture the variability and complexity of real-world deployment environments. The results might differ when models are deployed on different hardware, infrastructure or in varying operational contexts.
Also, the study focuses primarily on specific NLP tasks and may not generalize fully to other domains like vision or time series analysis. Additionally, while the study explores a range of system-related factors, it does not account for all possible variables that could influence inference energy, such as network latency or hardware-specific optimizations.


\section*{Ethical Considerations}
One of the main ethical issues in our experimentation was the substantial energy consumption and carbon emissions it produced. We perform 1024 inferences for 11 datasets over 10 models in several configurations, necessitating multiple repetitions of the inferences, along with several pilot experiments to finalize the experimental setup. This led to an approx total energy use of $3000$ kWh. To reduce our environmental impact, we limited our experiments to only 1024 test examples sampled from the datasets. 
We hope that the insights from this study will lead the community to a much larger reduction of the energy consuption of LLMs.


\section*{Acknowledgments}

The research is partially supported by a research grant from Accenture Corporation.
Paramita Koley is supported by SERB NPDF Fellowship.


\bibliography{custom}

\newpage 

\appendix

\section{Additional Literature Survey}
\label{sec:appen-literature}

\textbf{Sustainable Large Language Models:}
Schwartz et al. ~\cite{schwartz2020green} discuss the growing compute cost of deep learning research and advocate for making efficiency an evaluation criterion alongside accuracy and related measures with a focus on making AI both greener and more inclusive. 
Lacoste et al.~\cite{lacoste2019quantifying} consider various factors like energy grid, energy draw of server, make and model of training hardware to assess the environmental impact of machine learning algorithms. 
%
Following that trend, recent literature focuses on various alternatives to reduce the inference energy of large language models. Among the black-box approaches, Li et al~\cite{li2024toward} append generation directives to user prompts for carbon-friendly LLM inferences. \cite{mcdonald2022great} focus on techniques to measure energy usage and propose various hardware and datacenter-oriented settings that can be tuned to reduce energy consumption for training and inference for language models. Frugal GPT~\cite{chen2023frugalgpt} explores strategies like prompt adaptation, LLM cascade, and LLM approximation for reducing the inference cost for a large set of queries. 
On the contrary, white-box approaches include 
speculative decoding~\cite{leviathan2023fast}, speculative sampling~\cite{chen2023accelerating}, prunning~\cite{kurtic2024ziplm}, embedding recycling~\cite{saad2022embedding}, quantization~\cite{bai2022towards,frantar2022gptq,xiao2023smoothquant}, and many more.

\textbf{Tools for measuring energy impact:}
Researchers propose various tools for tracking the realtime energy consumption and carbon
emissions during model training and inferences. These tools include CodeCarbon~\cite{codecarbon}, CarbonTracker~\cite{anthony2020carbontracker}, Experiment impact tracker~\cite{henderson2020towards}, EnergyScope~\cite{limpens2019energyscope}, etc. 
Green Algorithms~\cite{lannelongue2021green} is another online tool, enabling a user to estimate and report the carbon footprint of their computation. Eco2AI is another open-source package to help data scientists and researchers to track energy consumption and equivalent CO$_2$ emissions of their models in a straightforward way~\cite{budennyy2022eco2ai}. %
Carburacy~\cite{moro2023carburacy} proposes the first carbon-aware accuracy measure that captures both model effectiveness and eco-sustainability for generative transformer-based models~\cite{moro2023carburacy}. 
Researchers explore energy impact analysis in terms of carbon footprints of ML algorithms in various domains, namely differential privacy~\cite{naidu2021towards}, medical image analysis~\cite{selvan2022carbon}, etc. 

\textbf{Benchmarking energy tools:} Researchers benchmark the tools for measuring carbon footprints in various configurations for various deep learning based ML models. 
Cao et al~\cite{cao2020towards} compare energy returned by software-based energy measurements with hardware power meter (WhattsUPMeter) on various NLP models and report experiment impact tracker as not so accurate. 
Jay et al~\cite{jay2023experimental} qualitatively and experimentally compare several software-based power meters
against high-precision physical power meters while executing various intensive workloads, where they conclude that for measuring energy, Carbon Tracker, Code Carbon, Energy Scope, and Experiment Impact Tracker are suitable fits. However, Bouza et al~\cite{bouza2023estimate} establish that the energy value reported by CodeCarbon is closest to Wattmeter, followed by CarbonTracker, with more variability between infrastructures.

\textbf{Benchmarking LLMs:}  Existing works have attempted to study the inference energy of LLMs from various perspectives. \citet{everman2023evaluating} evaluated energy usage of GPT models on question-answering tasks, employing Software Carbon Intensity (SCI) released by green software. \citet{samsi2023words} provides a detailed account of energy usage during inference for question-answering tasks for Llama models on various GPU architectures. \citet{liu2022few} study the energy consumption trade-off for fine-tuning vs few-shot learning for both encoder-decoder and decoder-only models on SuperGLUE and RAFT tasks, measuring energy in FLOPS. These preliminary works mostly ignore the performance-energy tradeoff. 

In recent times, \citet{li2024toward} compare the effect of prompt directives on inference energy and performance for Llama-2 models for 3 question-answering tasks. \citet{luccioni2024power} offers a comparatively detailed account of inference energy usage while choosing LLMs and tasks from a diverse range, showing the dependency of energy with model size and architecture, task type, and output token length. While these works provide an initial account of inference energy usage of LLMs in different configurations, they are often limited by the number and diversity of the models and tasks. Our work is the first to provide a comprehensive benchmarking of energy consumption of LLMs during inference for a diverse range of configurations, i.e., variation of input and output token length (for both coarse and granular level), correlation with response time, model size and family, task complexity, batch size, quantization level, presence of targeted phrases, and energy directives in prompt for a more complete set of LLM and tasks.
%
Table~\ref{tab:existing-literature} presents an overview of existing approaches and their limitations.
\begin{table*}[ht]
    \centering
    \footnotesize
    \begin{tabular}{p{27mm}|p{28mm}|p{24mm}|p{64mm}}
    \toprule
    \textbf{Reference } & \textbf{LLM} & \textbf{Tasks/Datasets} & \textbf{Observations}\\
    \midrule
    \cite{everman2023evaluating}  & GPT-styled models (4) & 10 manual prompts &  Study on the energy-performance tradeoff of LLMs.\\
    \hline
    \cite{samsi2023words} &  LlaMA models (3) & QA tasks (2) & Study inference cost on diverse GPUs.\\
    \hline
    \cite{desislavov2021compute} & DNN-based NLP models (7) & GLUE (9) & Study model complexity vs inference cost.\\
    \hline
    \cite{liu2022few} & T5 models (3), GPT-styled models (3) & NLP tasks (9), RAFT & Study energy consumption of few-shot PEFT vs in-context learning.\\
    \hline
    \cite{li2024toward} & Llama models (2) & QA tasks (3) & Study effect of prompt directives on inference cost.\\
    \hline 
    \cite{luccioni2024power} & Flan-T5 models (4), BLOOMz models (4) & NLP + vision tasks (10) & Study inference energy vs model complexity, task type, output, etc.\\
    \hline 
    Our work & GPT styled models (6), Flan-T5 models (4) & NLP tasks (11) & Study inference cost vs input, output, response time, model size \& family, task complexity, quantization, batch size, targeted phrases\\
    \bottomrule
    \end{tabular}
    \vspace*{-2mm}
    \caption{Comparison of our approach with existing literature on benchmarking inference cost of LLMs}
    \label{tab:existing-literature}
    \vspace*{-2mm}
\end{table*}

\section{Model Descriptions}
\label{app:models}
Check Table~\ref{tab:models-app}
\begin{table*}[!ht]
    \centering
    \footnotesize
    \begin{tabular}{l|p{110mm}}
    \toprule
    Model & Model description link\\
    \midrule
             
        \textbf{Tiny-LLama}~(1.1B params) & https://huggingface.co/TinyLlama/TinyLlama-1.1B-Chat-v1.0\\
        \textbf{Phi-3-mini}~(3.8B params) & https://huggingface.co/microsoft/Phi-3-mini-4k-instruct \\
        \textbf{Mistral-7B}~(7.2B params) & https://huggingface.co/mistralai/Mistral-7B-Instruct-v0.2 \\
        \textbf{Llama-2-7B}~(6.7B params) & https://huggingface.co/meta-llama/Llama-2-7b-chat-hf \\
        \textbf{Llama-3-8B}~(8.0B params) & https://huggingface.co/meta-llama/Meta-Llama-3-8B-Instruct \\
        \textbf{Llama-2-13B}~(13B params) & https://huggingface.co/meta-llama/Llama-2-13b-chat-hf \\ 

        \midrule
        
        \textbf{Flan-T5-base}~(248M params) & https://huggingface.co/google/flan-t5-base \\
        \textbf{Flan-T5-large}~(783M params)& https://huggingface.co/google/flan-t5-large \\
        \textbf{Flan-T5-xl}~(2.8B params)   & https://huggingface.co/google/flan-t5-xl \\
        \textbf{Flan-T5-xxl}~(11B params)   & https://huggingface.co/google/flan-t5-xxl\\
    
    \bottomrule
    \end{tabular}
    \caption{Links to specific models versions we used in our experiments}
    \label{tab:models-app}
\end{table*}

\section{Dataset Examples and Metrics}
\label{app:tasks}
Check Table~\ref{tab:tasks-app}

\begin{table*}[ht]
    \centering
    \footnotesize
    \begin{tabular}{p{40mm}|p{16mm}|p{15mm}|p{70mm}}
    \toprule
    \textbf{Task} &   \textbf{Dataset} & \textbf{Metric} & \textbf{Input Prompt with query} \\
    \midrule
    Linguistic acceptability check & COLA (GLUE) & Macro-F1 & Answer in binary whether the given sentence is grammatically, semantically, and logically acceptable. \\
    Logical entailment & Mnli (GLUE) & Macro-F1 & Select the stance of the premise towards the hypothesis: Entailment~(0), Neutral~(1) or Contradiction~(2). \\
    Sentiment classification & SST2 (GLUE) & Macro-F1 & Classify the sentiment of the sentence as positive~(1) or negative~(0). \\
    
    \hline
    Contextual question answering & Boolq (SuperGLUE) & Macro-F1 & Read the passage and answer the question with True~(1) or False~(0). \\
    Causal reasoning & COPA (SuperGLUE) & Macro-F1 & Select Choice1~(0) or Choice2~(1) that is a cause/effect of a given premise. \\
    Entity Question answering & ReCoRD (SuperGLUE) & F1 & Read the passage and find the entity that replaces ``@placeholder'' inside the query. \\
    Extractive question answering & SQuAD v2 & F1 & Read the context and answer the question with a phrase from the context. \\

    \hline
    Document summary generation & CNN-DM & avgROUGE-1,2,L & Summarize a given news article. \\
    Dialogue summary generation & SAMSum & avgROUGE-1,2,L  & Summarize a given dialogue sequence. \\
    
    \hline
    3 class vaccine-stance classification & VAX-Stance & Macro-F1 & Classify into one of the following three vaccine stances: Pro-Vaccine, Anti-Vaccine or Neutral. \\
    12 class multi-label anti-vaccine concerns classification & CAVES & Macro-F1 & Classify into one or more of these anti-vax classes: \qquad 0: ineffective, 1: ingredients, 2: rushed \textbf{...} 11: side-effect. \\
    \bottomrule
    \end{tabular}
    \caption{List of tasks/datasets we experimented on along with input prompts/descriptions}
    \label{tab:tasks-app}
\end{table*}

\section{Scatter Plots of batch-wise energy tracking}
\label{app:scatter}

\begin{figure*}[!ht]
    \centering
    
    \includegraphics[height=4.2mm,trim={86mm 141mm 40mm 2mm},clip]{Plots_main/scatter_model/flan-t5-base-energy_cc_vs_energy_ct.pdf}
    \vspace*{-4mm}
    
    {\includegraphics[width=0.3\linewidth]{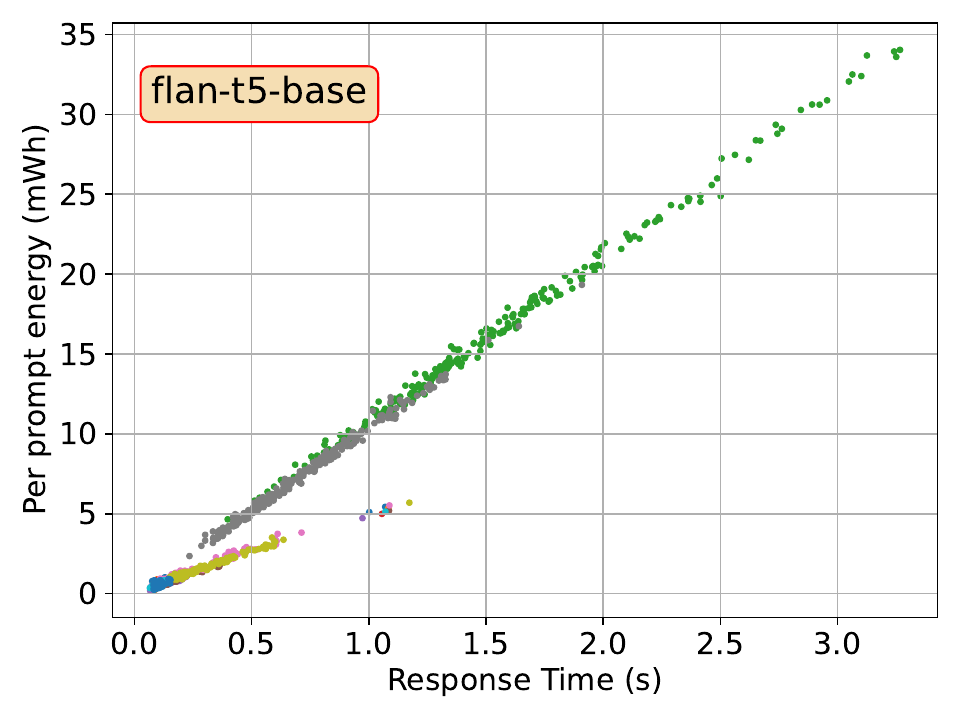}}
    {\includegraphics[width=0.3\linewidth]{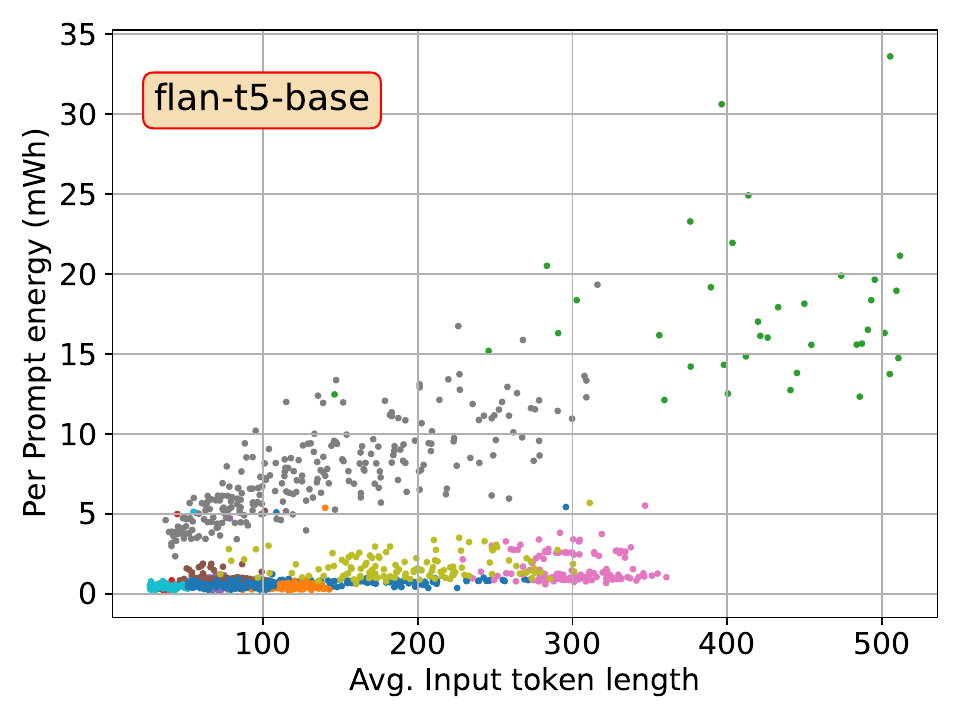}}
    {\includegraphics[width=0.3\linewidth]{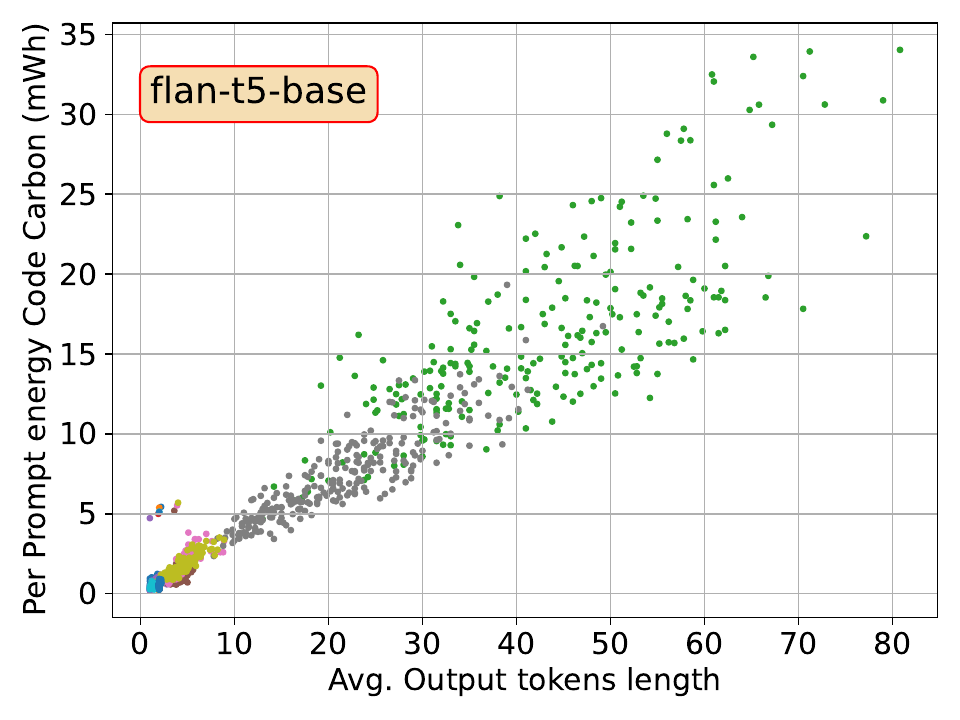}}

    {\includegraphics[width=0.3\linewidth]{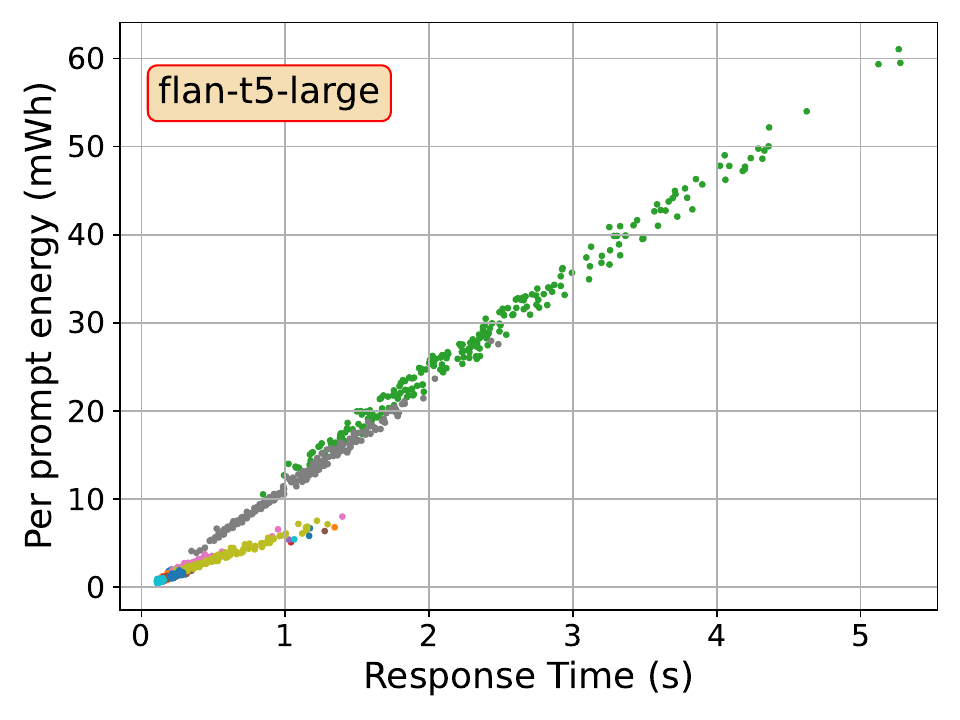}}
    {\includegraphics[width=0.3\linewidth]{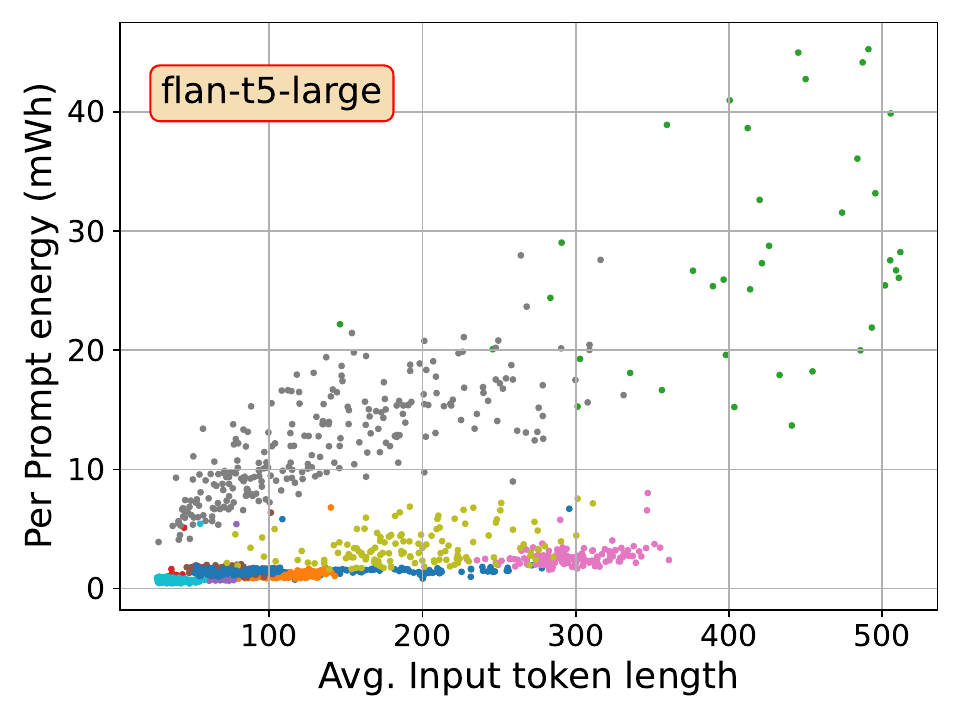}}
    {\includegraphics[width=0.3\linewidth]{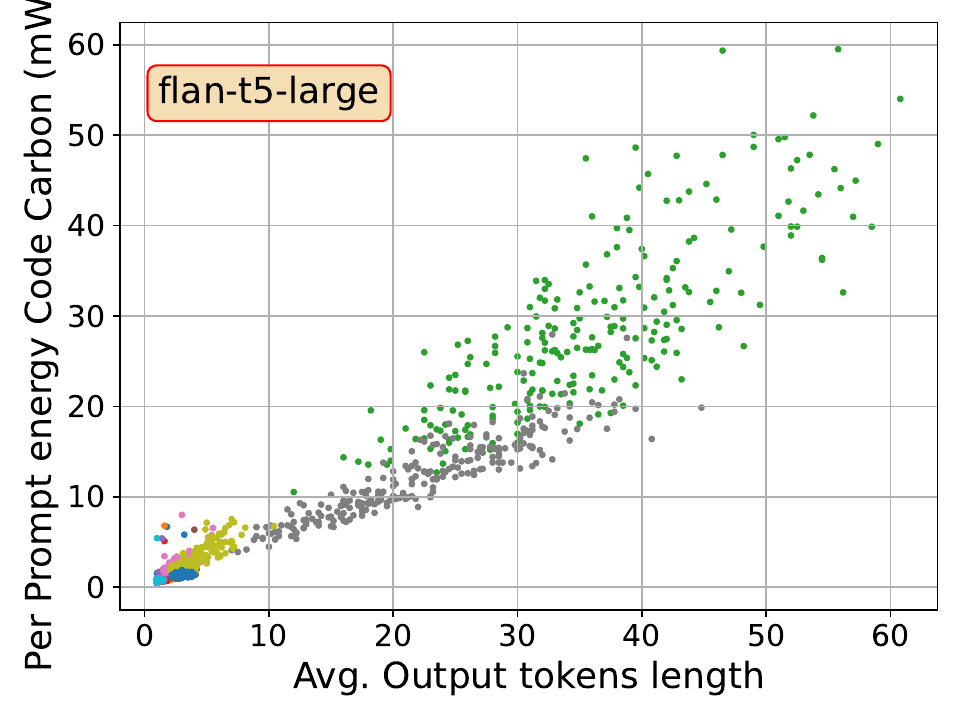}}

    {\includegraphics[width=0.3\linewidth]{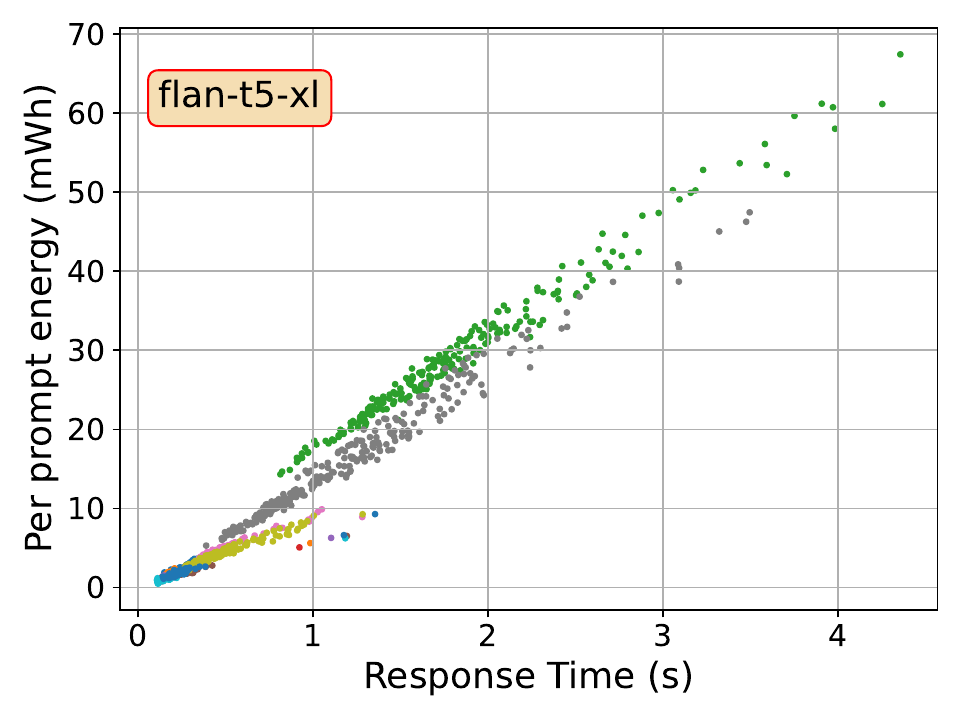}}
    {\includegraphics[width=0.3\linewidth]{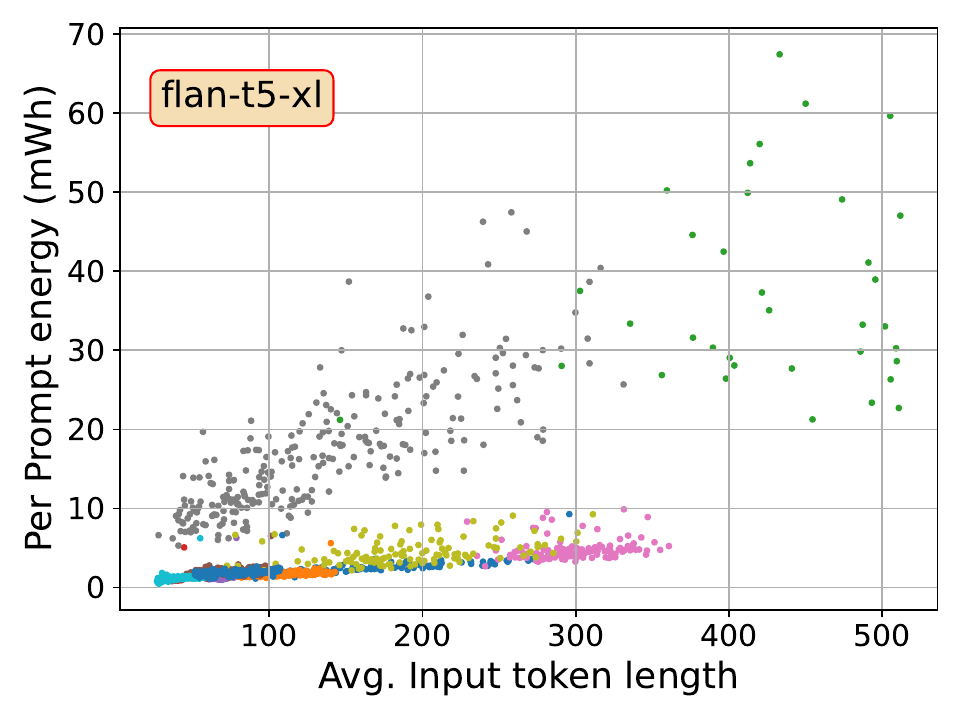}}
    {\includegraphics[width=0.3\linewidth]{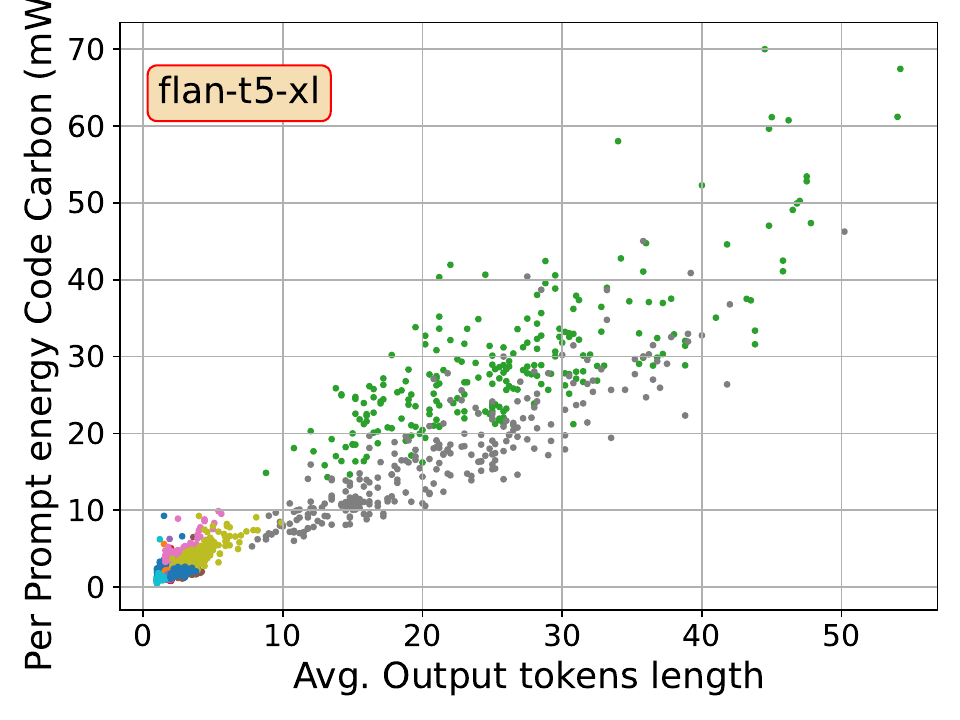}}

    {\includegraphics[width=0.3\linewidth]{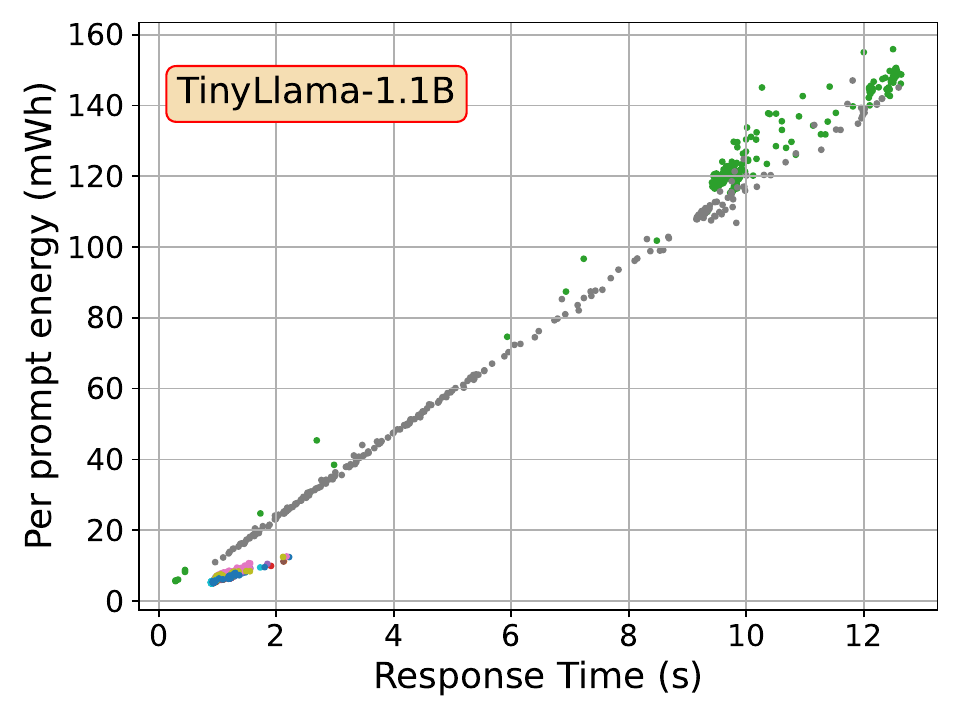}}
    {\includegraphics[width=0.3\linewidth]{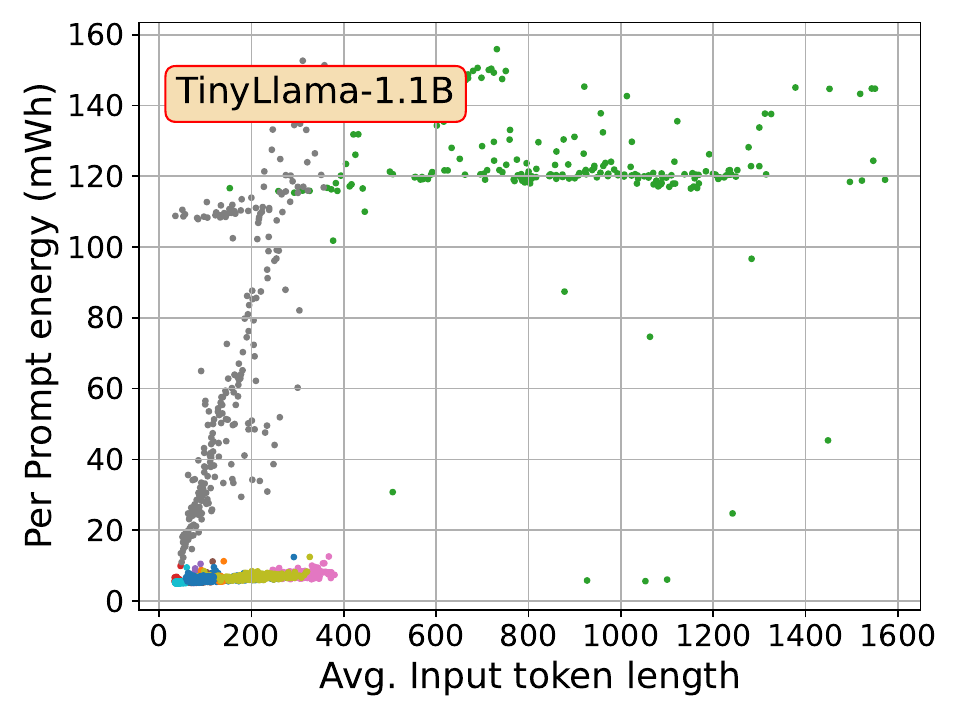}}
    {\includegraphics[width=0.3\linewidth]{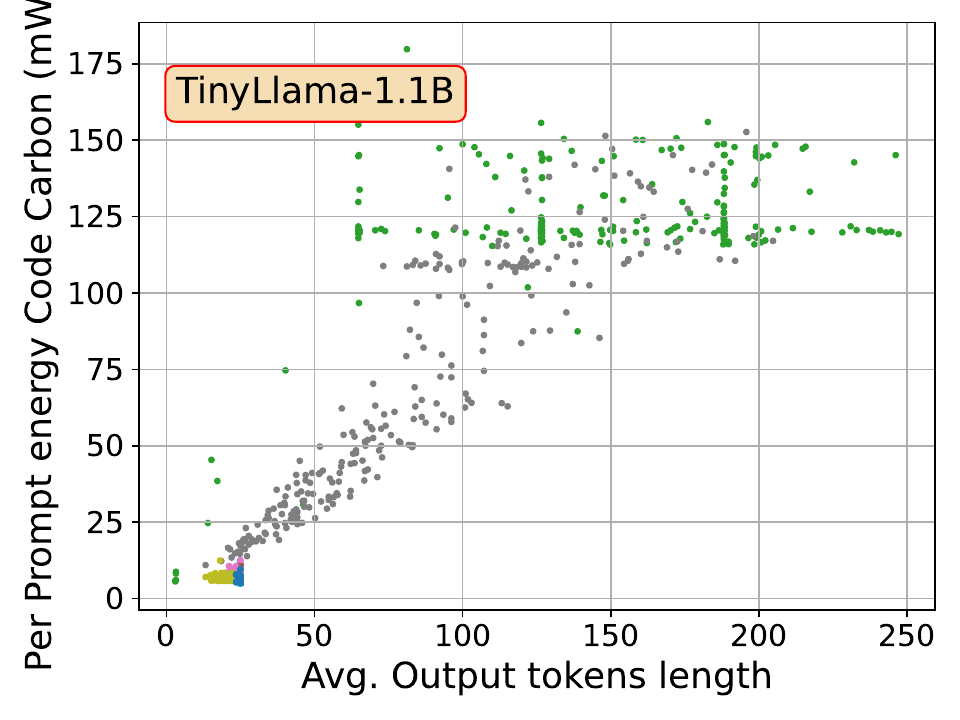}}


    {\includegraphics[width=0.3\linewidth]{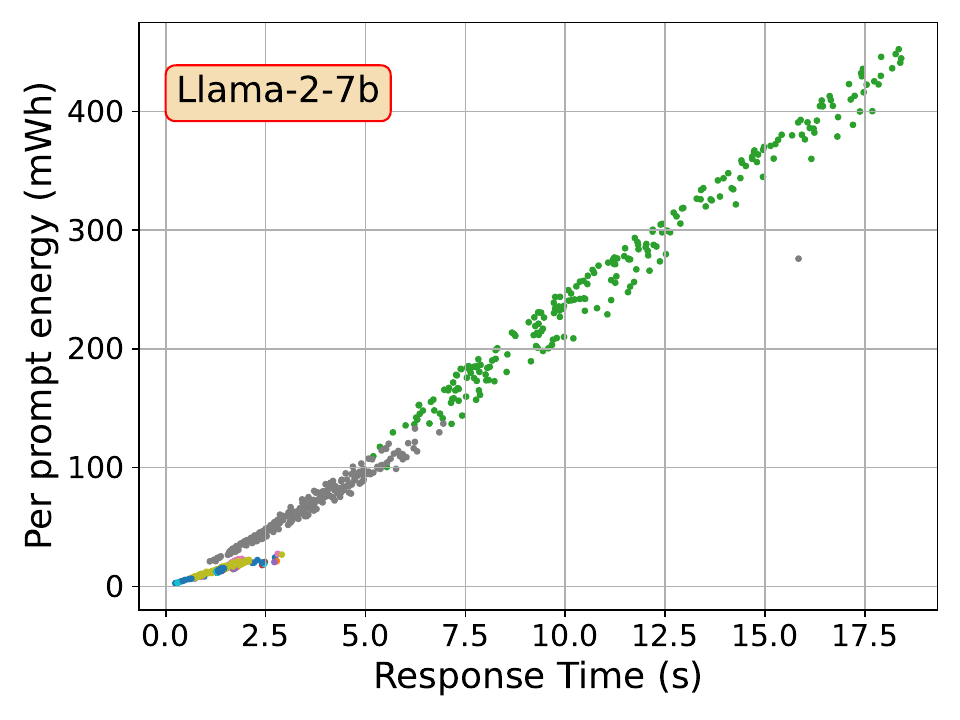}}
    {\includegraphics[width=0.3\linewidth]{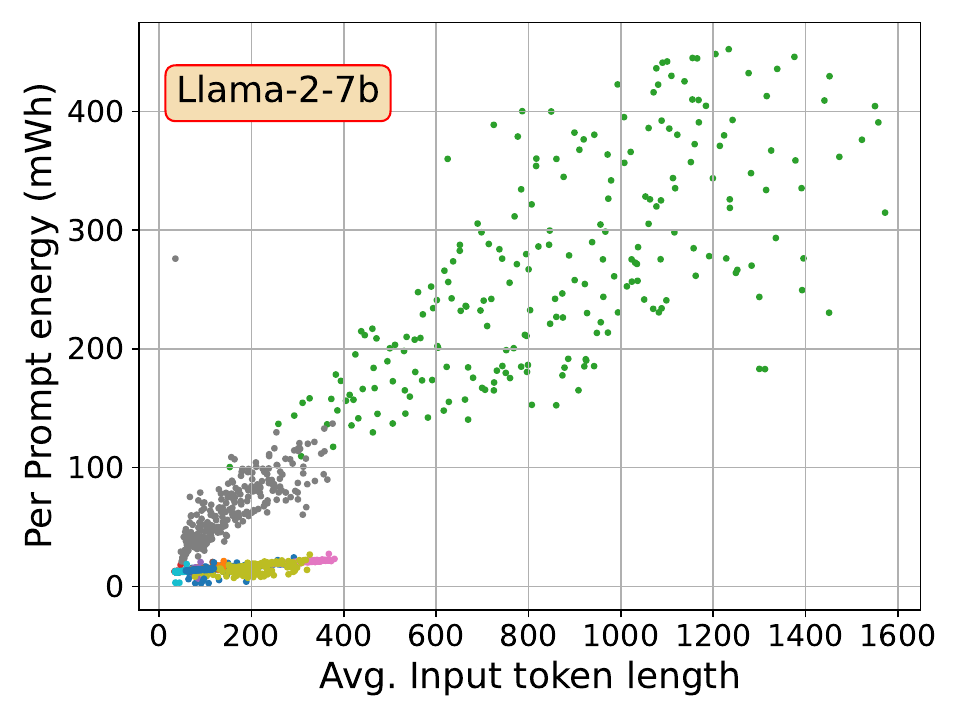}}
    {\includegraphics[width=0.3\linewidth]{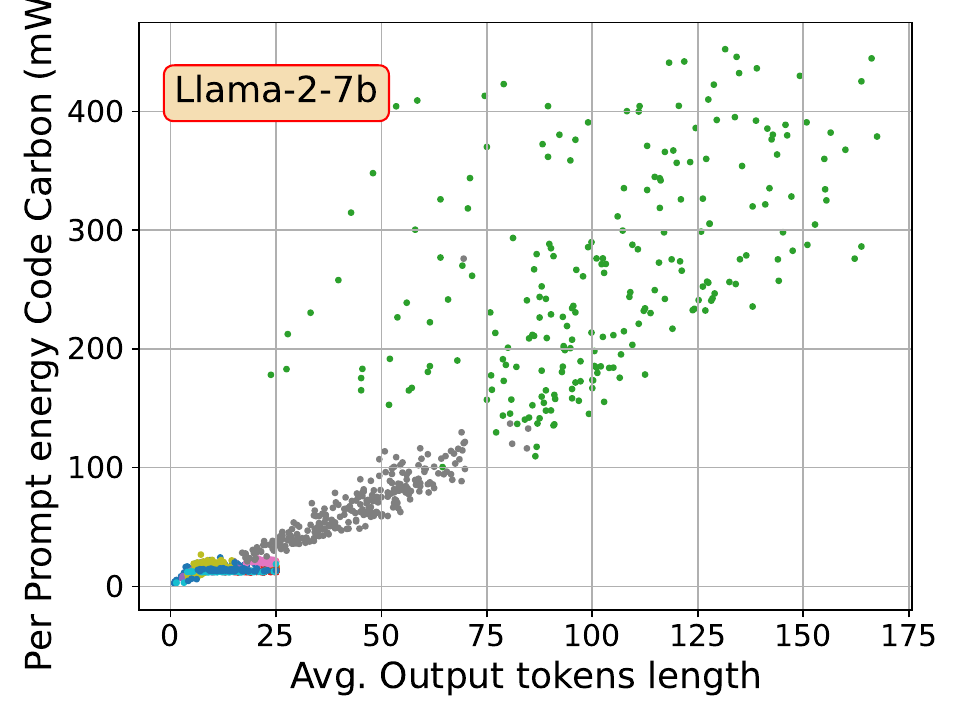}}

    {\includegraphics[width=0.3\linewidth]{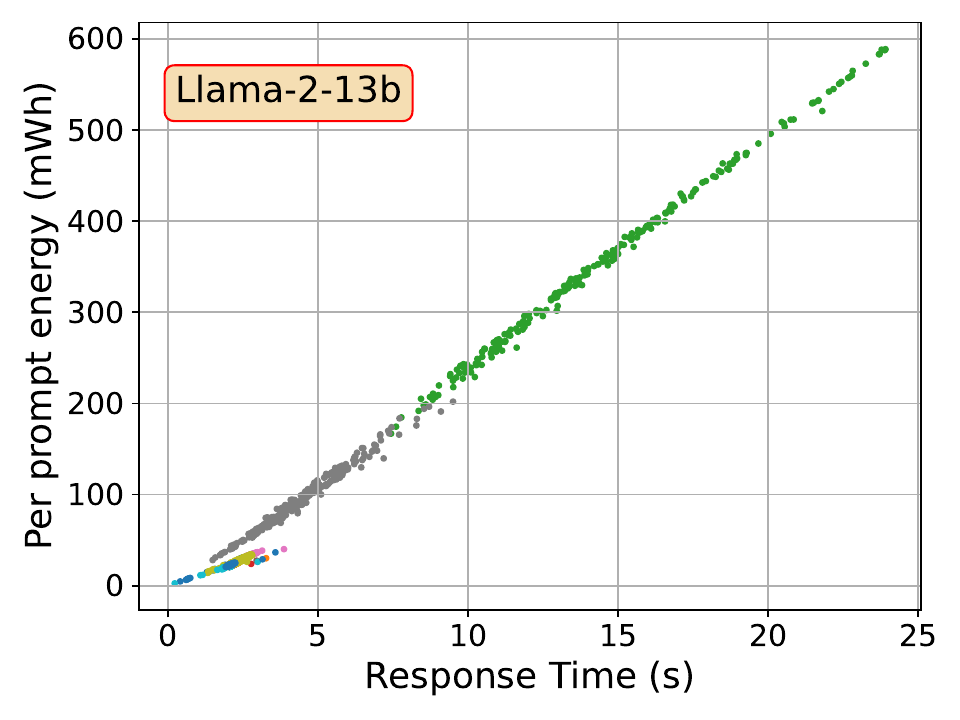}}
    {\includegraphics[width=0.3\linewidth]{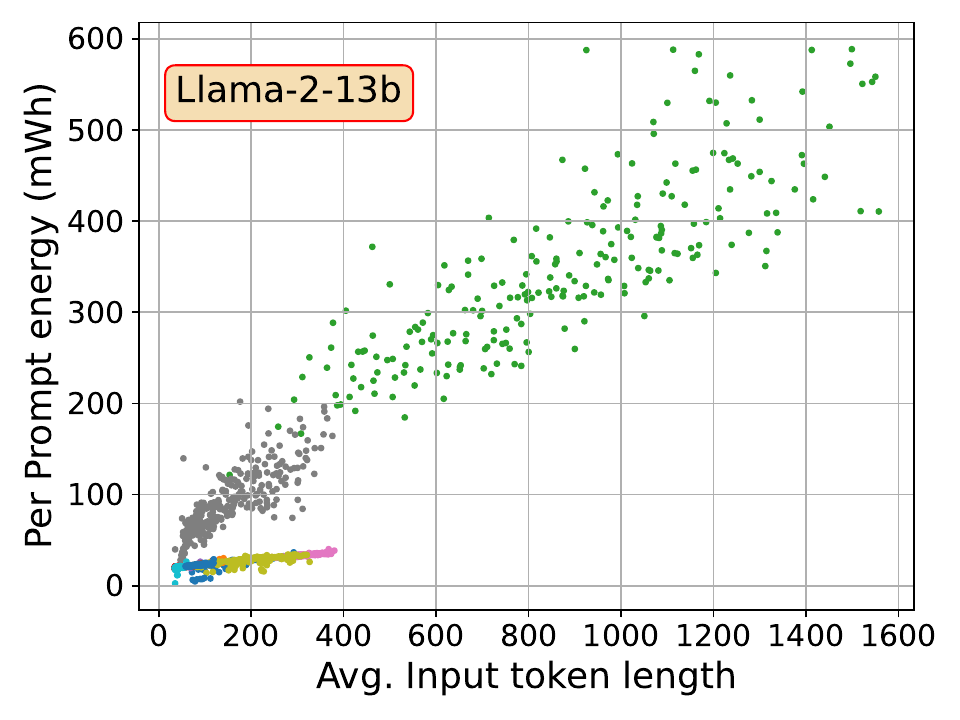}}
    {\includegraphics[width=0.3\linewidth]{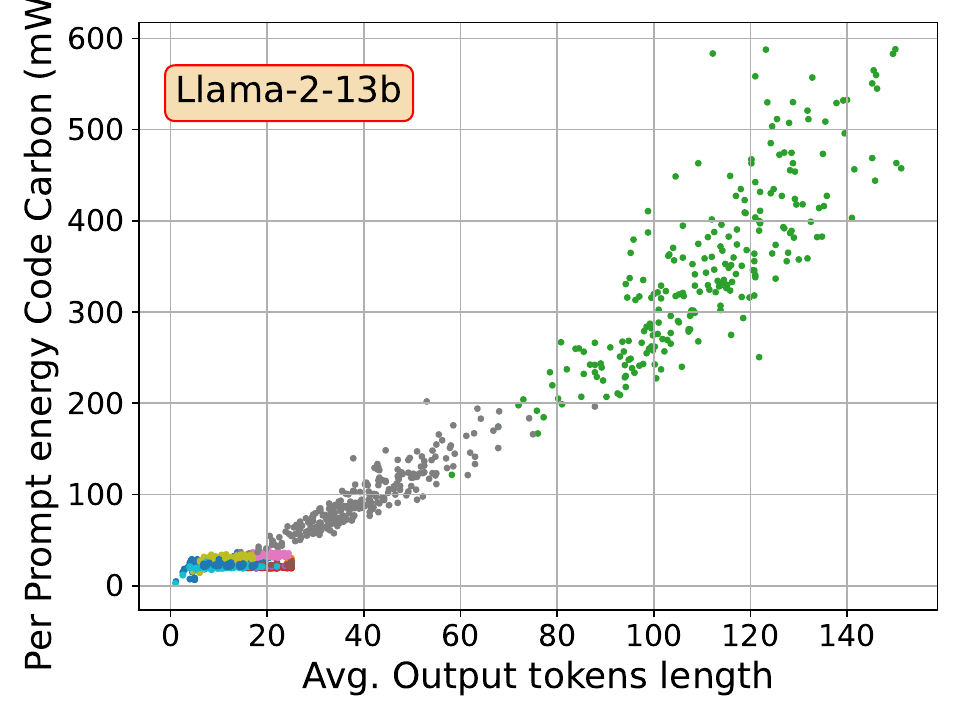}}

    \caption{Average per-sample inference energy vs average per-sample response time, input and output-token length across all datasets for different models where points in the image correspond to individual batches of different datasets.}
    \label{fig:scatter-app}
\end{figure*}

Check Figure~\ref{fig:scatter-app}

\section{Original Accuracy Metrics of individual datasets}
\label{app:scores}
Check Table~\ref{tab:metrics-app}

\begin{table*}[!ht]
    \begin{center}
    \footnotesize
    
        \begin{tabular}{|c||c|c|c|c||c|c|c|c|c|c|}
        \hline
        Dataset & flan-t5 & flan-t5 & flan-t5 & flan-t5 & TinyLlama & Phi-3 & Mistral & Llama-2 &  Llama-3 & Llama-2 \\
                                & base    & large    & xl      & xxl    & 1.1B      & mini  & 7B      & 7B      & 8B & 13B\\
        \hline \hline
        cola & 23.5 & 68.8 & 31.2 & 24.9 & 22.1 & 44.1 & 54.1 & 22.1 & 55.6 & 30.3\\ 
        mnli & 54.2 & 88.0 & 79.4 & 87.6 & 22.6 & 24.6 & 46.1 & 28.5 & 50.6 & 41.3\\ 
        sst2 & 33.0 & 74.5 & 32.7 & 40.2 & 47.4 & 51.8 & 75.1 & 48.6 & 71.1 & 57.7\\ 

        \hline
        
        boolq & 71.3 & 86.4 & 91.6 & 88.5 & 45.5 & 46.4 & 74.7 & 61.4 & 65.2 & 64.5\\ 
        copa & 33.3 & 41.9 & 26.0 & 42.7 & 36.7 & 64.9 & 60.3 & 53.2 & 73.7 & 56.1\\ 
        squad & 57.2 & 59.5 & 59.5 & 58.4 & 18.3 & 20.2 & 31.0 & 47.6 & 16.9 & 44.1\\ 

        \hline

        cnndm & 21.4 & 20.8 & 16.5 & 16.2 & 12.7 & 18.4 & 21.7 & 18.9 & 19.6 & 22.9\\ 
        samsum & 40.0 & 44.6 & 46.1 & 45.3 & 21.9 & 16.0 & 25.8 & 28.4 & 22.1 & 29.3\\ 

        \hline
        
        caves & 11.9 & 30.0 & 37.0 & 38.9 & 4.8 & 24.3 & 34.5 & 12.5 & 28.2 & 20.8\\ 
        vax & 20.3 & 53.0 & 52.1 & 54.6 & 23.0 & 47.9 & 52.7 & 50.2 & 52.5 & 54.2\\ 
        \hline
        \end{tabular}
        \caption{Original average ROUGE/F1 metrics for LLM inferences averaged across all datasets. 
        }
        \label{tab:metrics-app}
    \end{center}
\end{table*}

\section{Batch Size \& Quantization experiments}
\label{app:bs_quant}
\begin{figure*}[!ht]
    \centering
    
\subfloat[Per-sample inference energy averaged across all datasets when the batch size is varied. 
Overhead of using 4-bit precision can increase energy to almost $2\times$ for the same batch size. 
However, a 4-bit model in BS=256 takes only about $0.33\times$ the energy as default model in BS=64]{\includegraphics[height=50mm,trim={4mm 5mm 3mm 3mm},clip]{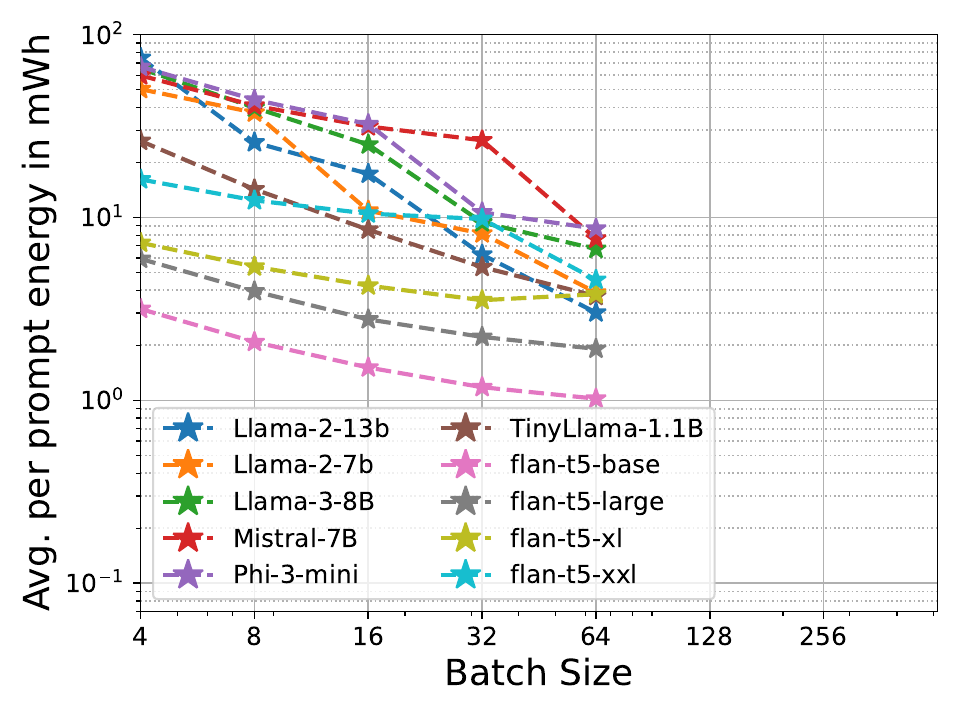}\label{fig:vary-batch-original}}
    \hfill

    \subfloat[Per-sample inference energy averaged across all datasets when the batch size is varied, on the A5000 GPU instead of A6000.]{\includegraphics[width=0.45\linewidth]{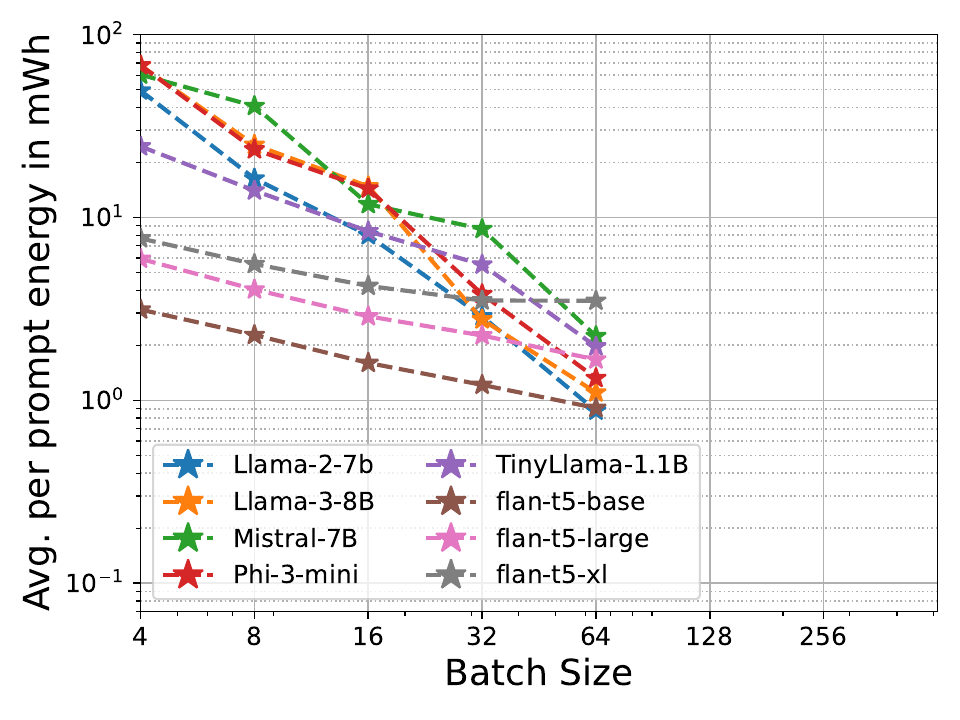}}
    \hfill
    \subfloat[Per-sample inference energy averaged across all datasets with 8-bit quantized models.]{\includegraphics[width=0.45\linewidth]{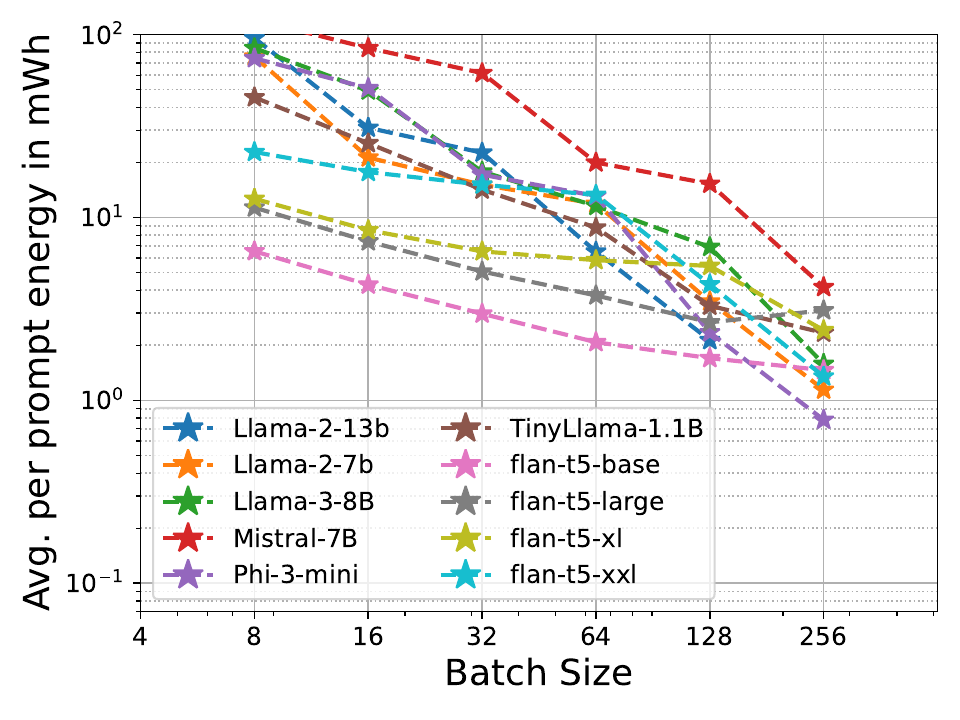}}    
    \caption{Additional Batch size experiments on the A5000 GPU, and using 8-bit quantization.}
    \label{fig:bs-quant-app}
\end{figure*}

Check Figure~\ref{fig:bs-quant-app}

\section{Testing on other systems}
\label{app:mgpu}

To verify the generalizability of our findings, we additionally ran limited experiments on two different systems. 

\noindent \textbf{System Descriptions:}
The first system contains an NVIDIA Tesla P100 GPU with 16GB VRAM paired with Intel Xeon Gold 6126 CPU and 128GB RAM. For the second system we used Kaggle~\footnote{www.kaggle.com/code} which contains an NVIDIA Tesla T4 GPU with 16GB VRAM, Intel Xeon CPU and 30GB RAM.

\begin{figure*}
    \centering
    \includegraphics[width=0.45\linewidth]{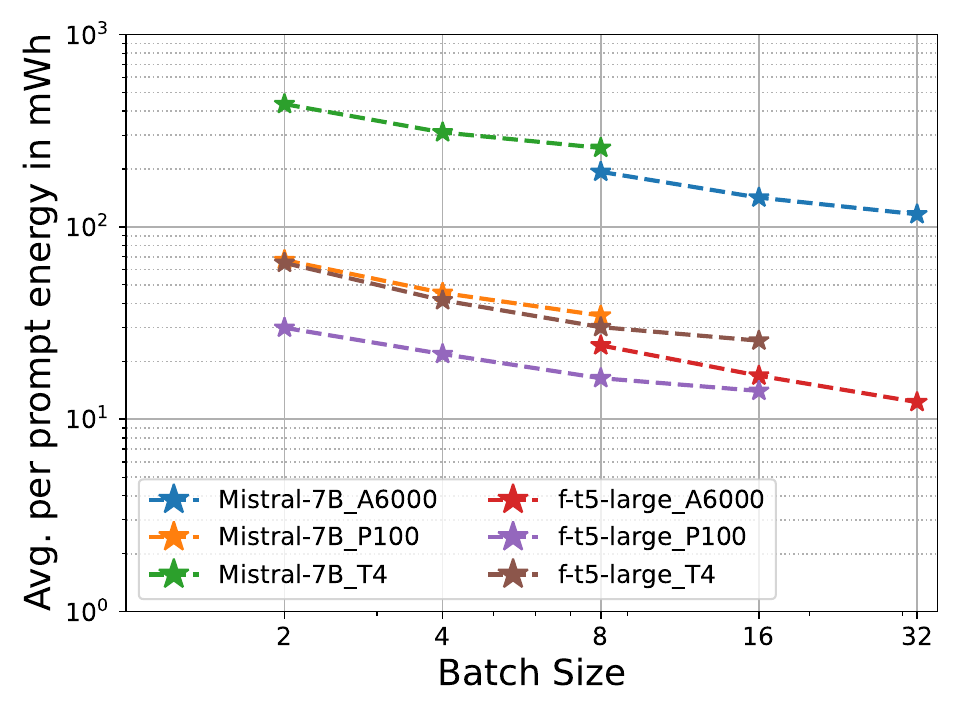}
    \caption{Energy consumption of running Mistral-7B and flan-T5-large on different systems (identified by their GPUs), averaged across 4 datasets. A6000 refers to the original setup.}
    \label{fig:mgpu}
\end{figure*}

\noindent \textbf{Experiments:}
Figure~\ref{fig:mgpu} shows the energy consumptions of 2 representative models from each family -- Mistral-7B and flan-t5-large on the different systems. The results are averaged over 4 representative datasets -- \boolq , \mnli, \cnndm and \squad.
We observe very similar trends for each model across different systems, even though the magnitude is different. We found that the system with P100 required less energy than our original setup, though with a limited GPU VRAM, which does not allow very high batch sizes. Interestingly, the system with T4 GPU consumes a lot more power even though it is a weaker system.  

We further hypothesize that modern improvements like flash-attention-2 will benefit newer GPU architectures and newer models like Phi-3 and Llama-3 more. However, the overall trends should still be similar.

\section{Energy proxy for black-box models}
\label{app:blackbox}
{
Our work demonstrates that for locally run open-source LLMs, inference time is a reliable proxy for estimating energy consumption, with significantly less overhead compared to using specialized energy measurement tools.
However, in the case of models accessed through online APIs (such as closed-source models), it is difficult to estimate the energy from just the response time due to a variety of factors, including network latency, number of concurrent user requests, type of batch scheduling/processing, etc. 
However, to date, the API providers do not provide any energy-related metrics; hence, there exists no way yet to estimate the energy for such black-box models. For the purpose of improving sustainability research, API providers should provide some energy metrics with the outputs.
In the absence of such estimates, inference time can be used to compare energy use of different black-box strategies. A workaround for variations in network latency could be to run experiments multiple times and averaging the time taken across all experiments. This will hopefully even out the effects of the varying factors such as network latency, when comparing different settings under the same hardware.
Though this would not completely eliminate the effect of network latency, as it is not a mean-zero random variable, and is influenced by a variety of complex, hardware-related factors, this is the most straightforward workaround for now.
In future work, it would be beneficial to develop a model that accounts for network latency variations to improve energy consumption estimation.
}

\end{document}